\RequirePackage{fix-cm}
\documentclass{svjour3}                     
\smartqed  
\usepackage{graphicx}
\usepackage{academicons}
 \journalname{}
\begin{document}

\title{Automated Defect Recognition of Castings defects using Neural Networks}


\author{A. García Pérez \and M. J. Gómez Silva \and A. de la Escalera Hueso}


\institute{
	A. García Pérez 
	\at
	ITP Aero, Alcobendas, Madrid, Spain \\
	Intelligent Systems Lab, Univ. Carlos III de Madrid, Spain \\
	Tel.: +34.91.2079108
	\email{alberto.garcia@itpaero.com}  \\
	\and
	M.J. Gómez Silva  and A. de la Escalera Hueso
	\at
	Intelligent Systems Lab, Univ. Carlos III de Madrid, Spain \\
}

\date{Received: date / Accepted: date}

\maketitle
\begin{abstract}
Industrial X-ray analysis is common in aerospace, automotive or nuclear industries where structural integrity of some parts needs to be guaranteed. However, the interpretation of radiographic images is sometimes difficult and may lead to two experts disagree on defect classification. The Automated Defect Recognition (ADR) system presented herein will reduce the analysis time and will also help reducing the subjective interpretation of the defects while increasing the reliability of the human inspector. Our Convolutional Neural Network (CNN) model achieves 94.2\% accuracy (mAP@IoU=50\%), which is considered as similar to expected human performance, when applied to an automotive aluminium castings dataset (GDXray), exceeding current state of the art for this dataset. On an industrial environment, its inference time is less than 400 ms per DICOM image, so it can be installed on production facilities with no impact on delivery time. In addition, an ablation study of the main hyper-parameters to optimise model accuracy from the initial baseline result of 75\% mAP up to 94.2\% mAP, was also conducted.
 
\keywords{Keywords: X-rays \and Castings defects\and Automated inspection \and ADR \and Convolutional Neural Network \and RetinaNet}
\end{abstract}

\section{Introduction}
\label{intro}
As part of the quality assurance process, industrial products are subject to numerous inspections such as visual, liquid penetrant, eddy current, ultrasonic or X-ray inspections. The latter technique allows to detect not only superficial defects but also internal defects in structural components, where any minor defect may lead to a potential crack in service operation or an unnoticed early failure. In recent years, the arrival of Digital Radiography (DR) permitted to acquire digital images directly from the sensor array. The good quality of such images allows to carry out metrology analysis, such as dimensional control, wall thickness measurement or process optimization amongst others. It also permits to inspect the interior of the part to detect flaws such as cavities, cracks, pores, shrinkages, inclusions etc. with an accuracy not seen before. 

In the particular case of industrial castings parts, the number and variety of defects can be significant in terms of type, texture and size. For instance, shrinkages are material fractures or bubble shaped voids created due to non-homogeneous regions during molten metal cool down process. Fractures are caused by mechanical stress due to thermal stress gradients, whilst voids are caused by the liquid metal failure to flow into the die. Some of the main shrinkage defects are sponge, filamentary and dendritic shrinkages and sometimes are very difficult to distinguish amongst them. Sponge shrinkages are usually located on mid-section of the casting, and these fractures are visually very similar to a sponge. On the other hand, filamentary shrinkage are produced by a network of continuous cracks, usually under a thick section of material, with fracture lines interconnected with various dimensions and densities. Finally, dendritic shrinkages are fractures, thinner and less dense than filamentary cracks, which are unconnected or with randomly distributed lines. 

\begin{figure}
	\centering{}
	\includegraphics[width=0.24\textwidth]{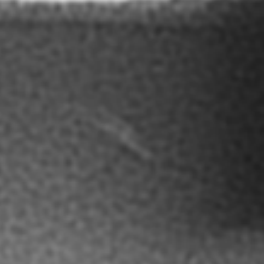}
	\includegraphics[width=0.24\textwidth]{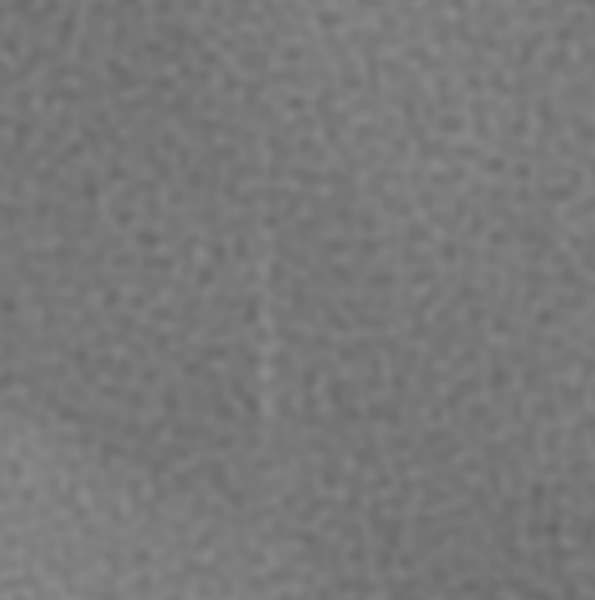}
	\includegraphics[width=0.24\textwidth]{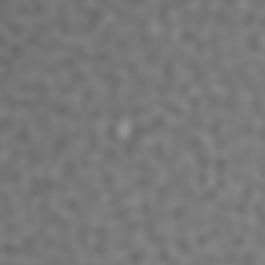}
	\includegraphics[width=0.24\textwidth]{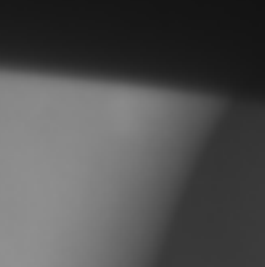}
	\caption{Samples of castings defects: ceramic inclusion, linear indication, porous, sponge shrinkage}
	\label{fig:castings-defects}
\end{figure}

In addition, castings parts can also have other defects produced, for instance, by trapped air and other gases, dissolved in the molten metal, causing porosity (or pinholes), either inside in the casting (internal defect) or on the surface. Sometimes, they appear in large numbers, creating the so called clusters. The presence of unwanted particles, such as ceramic particles coming the mould are known as inclusions and may lead to reduction on metal properties. Cavities are caused by turbulence of the molten fluid, which leads to the creation of internal or exposed depressions on the surface during the solidification process and usually they appear with angular edges. In addition, hot tear or cracks may also appear, caused by a poor mould design may lead to thermal tensile forces during the solidification process above the strength of the solidifying metal. These are only a few types of defects that can be found on industrial castings, which shows the complexity and the degree of expertise required by the human inspector to detect and classify such variety of defects.

In production, the analysis of the X-ray images is done by certified inspectors using expensive computer (medical) displays and under low light conditions, in order to capture the maximum contrast as possible to detect fine-grain defects. Usually, there are also limitations on the maximum time that the inspector can spend reviewing the images before a compulsory break is required to avoid visual tiredness and fatigue. On a single work turn, a human inspector may review a few hundreds of images in a process that can be monotonous and error prone. 

Hopefully, the use of Machine Learning (ML) techniques, in particular, Deep Learning (DL) methods have shown an impressive advance in the last decade to resolve Computer Vision (CV) problems while achieving or even exceeding human performance. One of the major Machine Learning models groups are Discriminative models, that discriminate between different kinds of data by calculating a probability on how likely a label is on the input instance. In particular, Convolutional Neural Networks (CNN) are now the industry standard to solve classification, object detection and segmentation problems. 

\begin{figure}
	\centering{}
	\includegraphics[width=0.32\textwidth]{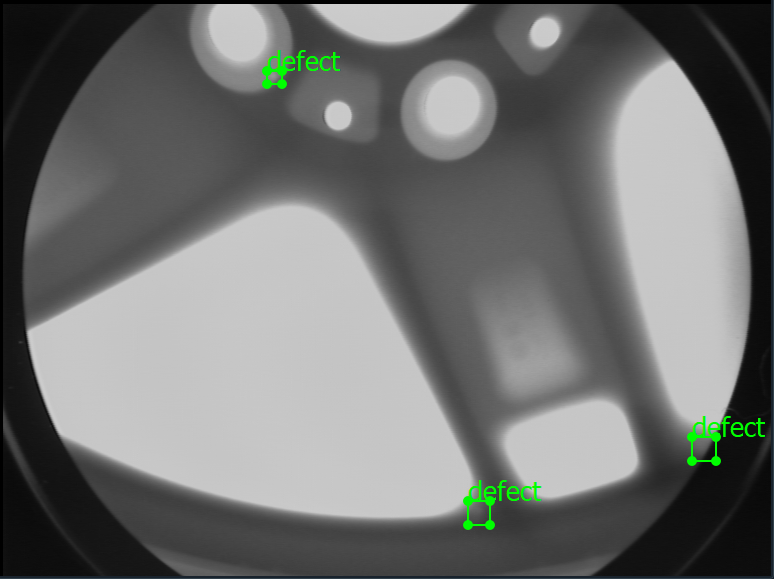}
	\includegraphics[width=0.32\textwidth]{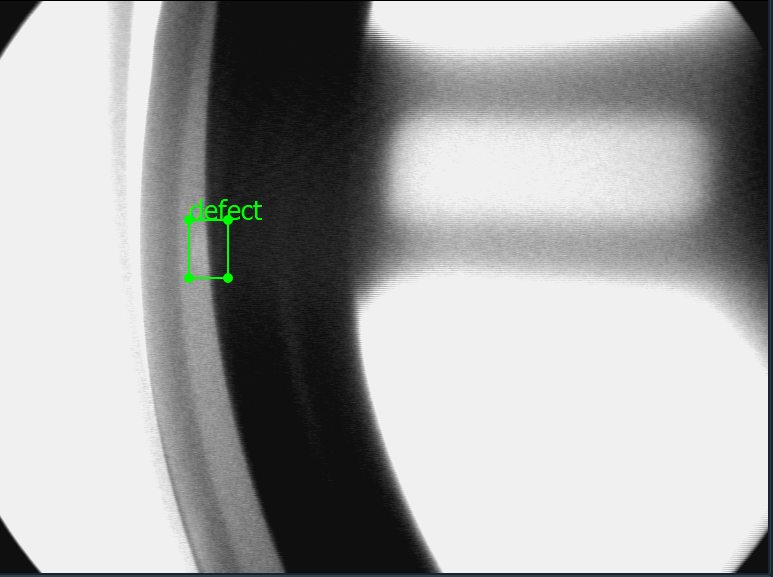}
	\includegraphics[width=0.32\textwidth]{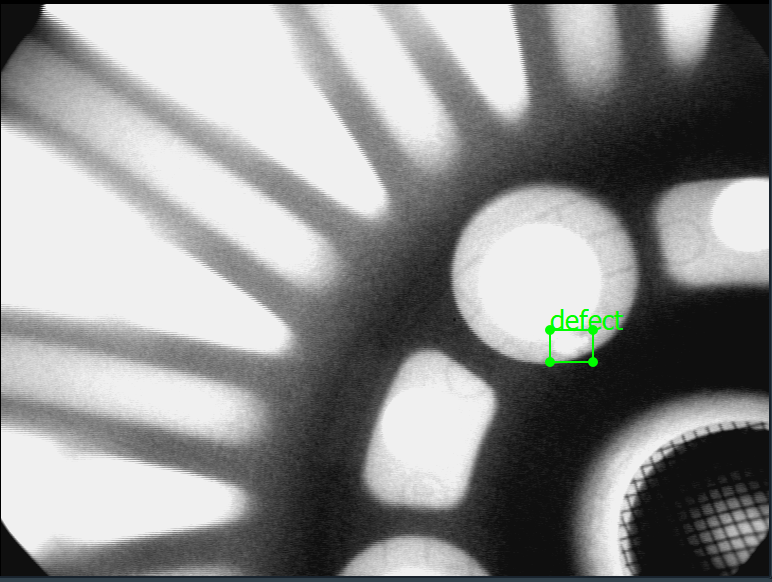}\\
	\includegraphics[width=0.32\textwidth]{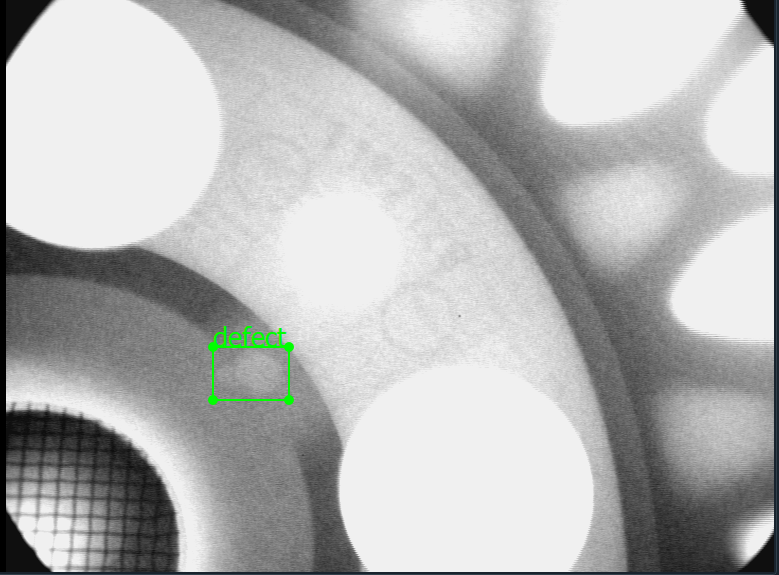}
	\includegraphics[width=0.32\textwidth]{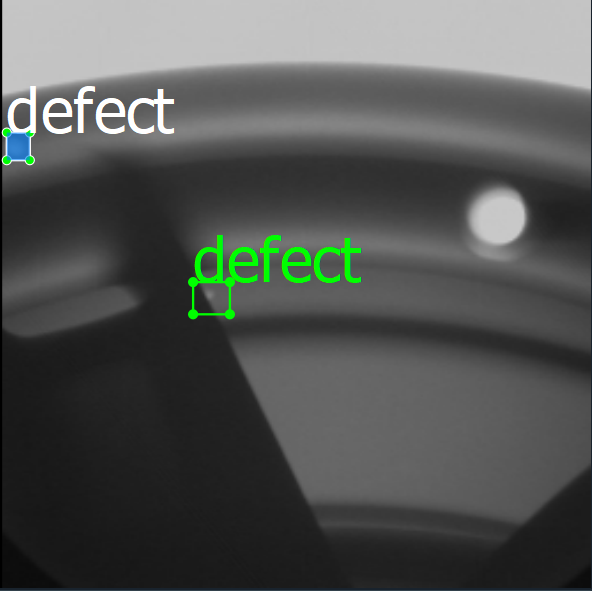}
	\includegraphics[width=0.32\textwidth]{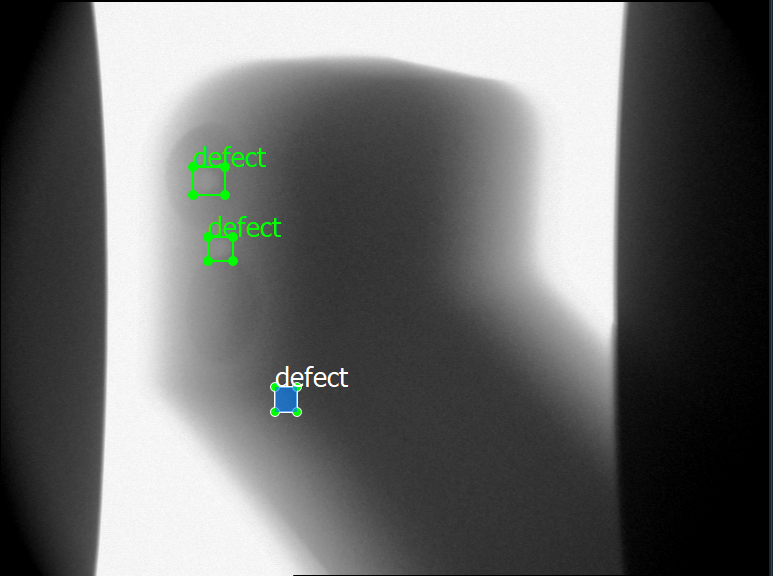}
	\caption{Sample images of GDXray Dataset \cite{meryGDXrayDatabaseXray2015a}}
	\label{fig:GDXray-samples}
\end{figure}

In this work, an Automated Defect Recognition (ADR) model is presented that exceeds current state of the art accuracy \cite{fergusonDetectionSegmentationManufacturing2018} and based on a CNN RetinaNet model \cite{linFocalLossDense2018} combined with a Feature Pyramid Network (FPN) \cite{linFeaturePyramidNetworks2017} in order to improve its performance at different defect scales. Original GDXray dataset images \cite{meryGDXrayDatabaseXray2015a} were also upscaled to improved detectability of smallest defects, without adding a computational penalty on our model. A thoughtful approach was also followed in order to explore the loss function topology to reach best local minima, showing that Single Stage Detector (SSD) models \cite{liuSSDSingleShot2016} can exceed the traditionally more accuracy two stage models such as Faster R-CNN \cite{renFasterRCNNRealTime2016} or Mask R-CNN  \cite{heMaskRCNN2018} if appropriate set of hyper-parameters are used. The proposed model also is relatively light and can be run on a conventional laptop with acceptance inference time. 

Section 2 covers a review on how automated inspection was approached in the past for industrial welding and castings defects. Section 3 describes the RetinaNet model used for the proposed Automated Defect Recognition system and how it resolves the learning process of the more difficult and small defects in the dataset, while also describes the main metrics and criteria employed for this object detection problem. Section 4 shows the experimental results and ablation analysis carried out in order to optimise model performance. Finally, section 5 presents the conclusions of our analysis.

\section{Related work}
\label{relatedwork}
Early attempts to create an ADR were based traditional computer vision techniques using manual or semiautomatic feature extraction as described in Boerner and Strecker \cite{boernerAutomatedXrayInspection1988}. Mery et al. \cite{meryPatternRecognitionAutomatic2003} extracted geometric features, Fourier descriptors, and gray value features such as mean value, mean gradient in the boundary, invariant moments, textural features or Discrete Fourier transforms to create a new contrast feature that allowed to create a correlation between feature and class of defects. Similarly, Li et al. \cite{liImprovingAutomaticDetection2006} proposed a wavelet technique where high-frequency regions of the image where compared against low frequency background to detect the defects. However, all these approached often failed at distinguishing between defects and part local geometry such as holes, edges or direct view of the X-ray radiation to the sensitive film material.

On a second stage, a more complex approach was followed based on fuzzy models. Hernández et al. \cite{hernandezNeuroFuzzyMethodAutomated2004} applied also feature extraction and data compression as an input to a neuro-fuzzy method to classify aluminium castings patterns which achieved a ROC value of 0.9976. Tang et al. \cite{tangApplicationNewImage2009} presented a fuzzy exponential entropy model for object and background segmentation based on predefined parameters and gray level distribution of the image. 

On a third stage, the approach to resolve the problem finally focused in the use of Artificial Neural Networks (ANN). Zapata et al.\cite{zapataAutomaticInspectionSystem2012} presented an automatic inspection system for welding radiographic images. On a first step, image preprocess was used to improve contrast and reduce noise in order to apply feature extraction to identify the welding region and defects segmentation. On a latter stage, such features were subject to a Principal Component Analysis (PCA) as a way to reduce dimensionality and eliminate redundancy and introduced on an ANN where they introduced a regularisation process to improve the performance of the classification stage. Mery et al. \cite{meryAutomaticDefectRecognition2017} explored several computer vision techniques on GDXray dataset \cite{meryGDXrayDatabaseXray2015a} including Local Binary Patterns (LBP) combined with linear Support Vector Machines (SVM) as classifier. 

However, in practice, classifying an image with a single label is not enough, and more information is desirable as how many defects are there in an radiography. This is the case for object detection problems where the objective is to define bounding boxes pointing to an object plus giving a label to such defect (a classification problem). Hence, object detection is a more difficult task not only because of the bounding box location, suppression of repeated bounding boxes etc., but also the detection of small objects or objects subject to occlusion, where some important features of the object may be not visible but we still want to detect it. Using this approach, Fergusson et al. \cite{fergusonDetectionSegmentationManufacturing2018} proposed a Mask Region based Convolutional Neural Network (CNN) that could be used to simultaneously perform defect detection and pixel segmentation based on GDXRay castings and welding database (GRIMA - Group of Machine Intelligence, Mery et al. \cite{meryGDXrayDatabaseXray2015a}). The Mask R-CNN model \cite{heMaskRCNN2018} was based on a ResNet-101 backbone \cite{heDeepResidualLearning2015} to create region proposals and on a Faster R-CNN model \cite{renFasterRCNNRealTime2016} for the Region Based Detector (RBD) that finally classifies the casting defects and identified the associated bounding box coordinates. By using only the object detection part of the model, Ferguson achieved 0.931 mAP but could be increased up to 0.957 mAP using segmentation techniques which involved a complex and time-consuming process of creation of the exact pixels that define the Ground Truth (GT) defect. 

Subsequently, Mery et al.\cite{meryAluminumCastingInspection2020} also proposed to use a CNN model where additional synthetic defects were introduced to improve model accuracy, either by means of 3D ellipsoidal models or by using Generative Adversarial Networks (GAN) \cite{goodfellowGenerativeAdversarialNetworks2014}. By using a sliding-window methodology, they achieved 0.71 mAP.

Our approach achieves 0.942 mAP using an object detection model, that outperforms current state of the art accuracy defined by Ferguson et al. \cite{fergusonDetectionSegmentationManufacturing2018} without using segmentation techniques. Our Automated Defect Recognition (ADR) is based on a CNN RetinaNet model \cite{linFocalLossDense2018} combined with a Feature Pyramid Network (FPN) \cite{linFeaturePyramidNetworks2017} and an upscaling of the input images in order to improve its detection performance at different defect scales. 

\section{Object detection methodology}
\label{sec:objectdtection}
\begin{figure}
	\centering{}\includegraphics[scale=0.25]{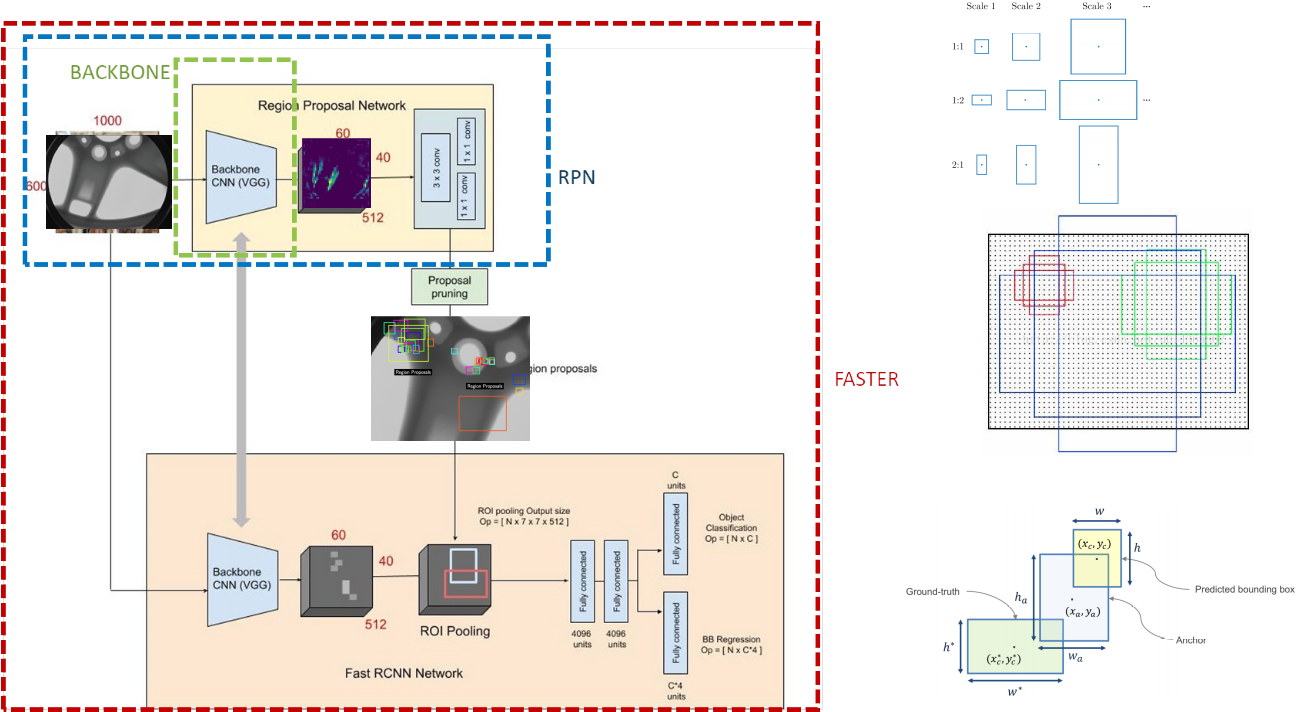}\caption{Faster RCNN architecture}
\end{figure}

Image object detection is a very computer intensive task as it usually involves searching in a large area for multiple objects, with different sizes and aspect ratios, object bounding box location, suppression of repeated bounding boxes etc. It is also a very challenging the detection of small objects or objects subject to occlusion, where some important features of the object may be not visible but we still want to detect it. In addition, model latency is a key reference parameter to compare model performance, in addition to some metrics as mean Average Precision (mAP) that will be described in subsequent sections. Model latency is key in applications such as robotics or self-driving cars in addition to model size, as in most of these applications the inference needs to be done using modest on-board computer resources, so the lower number of weights in a model, the better.
Two different approaches are available nowadays to resolve the object detection problems based on Single Stage Detectors (SSD) or in two steps detectors. However, both methodologies use the same pipeline: given an input image, they generate proposals or areas where an object could be found. These proposals can be generated by background subtraction, using sliding windows, pyramid of images or by selective search (based on shape, texture etc.). Then they classify these regions and, finally, a Non Maximum Suppression (NMS) algorithm is applied to eliminate multiple detections of the same object, i.e. several bounding boxes overlapping each other around the detected object. This algorithm defines the best bounding box and remove the rest.

Two stage detectors were considered, until recent improvement on SSDs, as more accurate and flexible models by using a two steps approach as follows:
\begin{itemize}
\item{A Proposal Generator, that extracts all the bounding boxes in the image that may contain an object. This first stage reduces the number of candidate object locations to a relatively small number (typically 2000 potential objects). Its main purpose is to remove or filter the background samples (with no rich information) from the potential object locations having rich information, such as edges, corners, sudden changes in texture, etc.}

\item{A Refine Stage, that classifies and redefines the bounding box for each candidate object location.}
\end{itemize}

In the other hand, SSDs are traditionally considered as simpler models and in some way more efficient due to the fact of using predefined anchors, in such way that they have recently exceeded the precision of two stage models on the latest architectures that have appear. Our proposal is based on a SSD model, named RetinaNet, that uses a ResNet-50 backbone with FPN as described in the following sections.

\subsection{ResNet backbone}
\label{sec:resnet}

\begin{figure}
	\centering{}
	\includegraphics[scale=0.5]{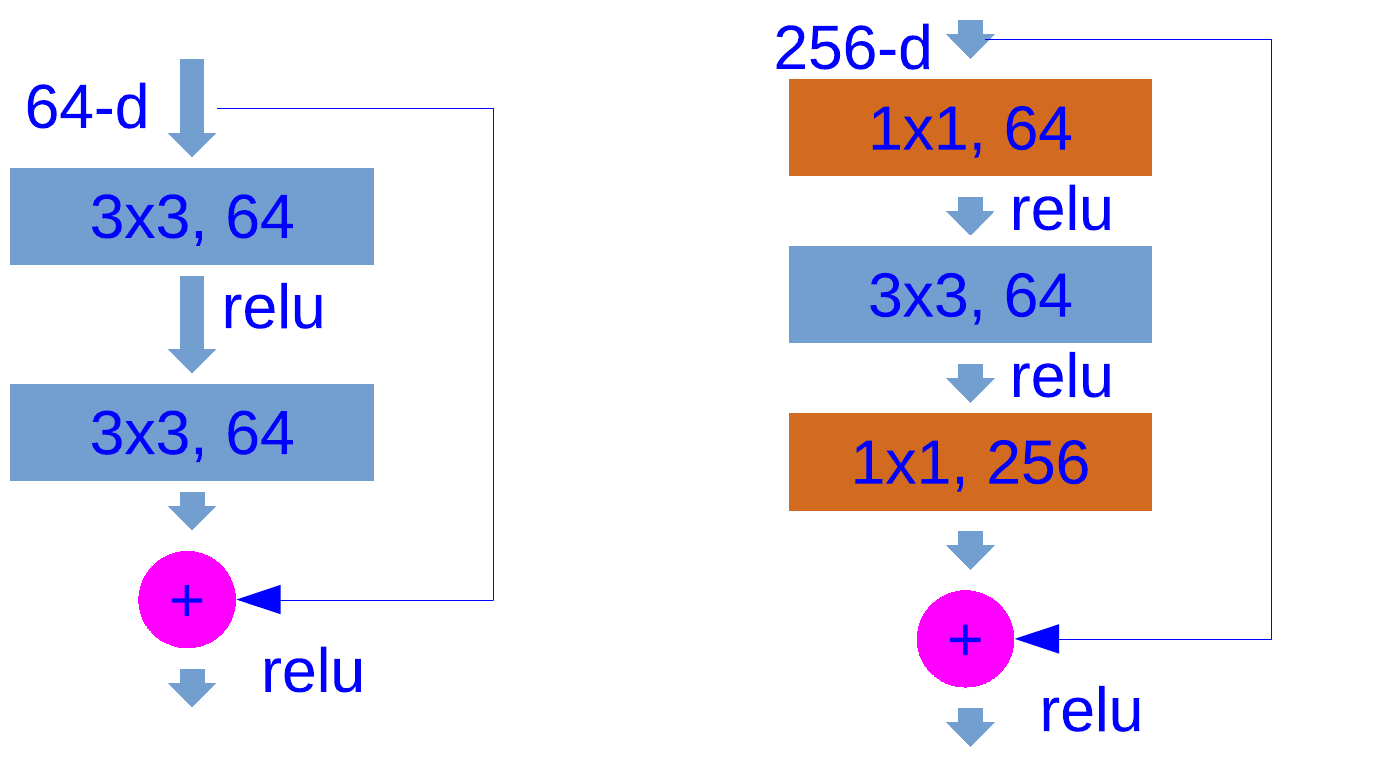}
	\caption{Residual module defined in ResNet backbone. Left module corresponds to ResNet-34, whilst right module corresponds to ResNet-50/101/152.\cite{heDeepResidualLearning2015}}
	\label{fig:resnet}
\end{figure}

The deeper the convolutional network is, the more important is the vanishing gradient problem. To prevent this problem, He et al. \cite{heDeepResidualLearning2015} introduced the ResNet architecture which introduces a residual module as shown in Figure \ref{fig:resnet}. Instead of using four branches, as does the Inception module, only uses two branches: left and right branches. The left branch consists on a series of convolutions (1x1, 3x3), activations and batch normalizations. On the other hand, the right branch is a “linear shortcut”, which connects the module input (or output from previous layer) with the bottom of the module where it is added to the output of the left branch. This addition, known as an identity mapping, turns out the network to learn faster and with larger learning rates, allowing to train networks that are substantially deeper than previous architectures, reducing the problem of the vanishing gradient problem.

In addition, ResNet does not rely on max pooling operations to reduce feature map size and relies on convolutions with strides ${>}$ 1. ResNet has also five stages where all layers on the same stage share the same convolution type, except for the first layer which is used to perform a down-sampling.

\subsection{Feature Pyramid Network} \label{sec:fpn}
\begin{figure}
	\centering{}
	\includegraphics[scale=0.8]{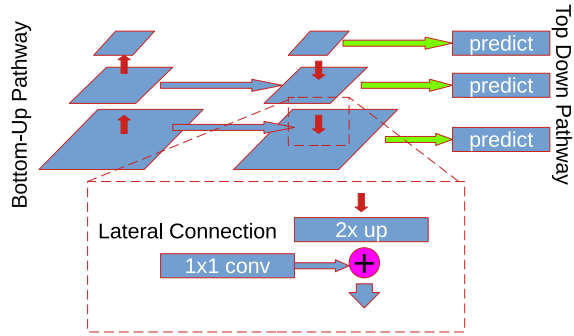}\caption{Example of Feature Pyramid Network showing the lateral and top-down connections \cite{linFeaturePyramidNetworks2017}}
	\label{fig:FPN}
\end{figure}
Each layer on a CNN produces a feature map with decreasing spatial resolution but with stronger semantic features as the convolution progress to the end of the network. Lin et al. \cite{linFeaturePyramidNetworks2017} realised that some key information was abandoned on each ConvNet stage, simply because of the reduction in the feature map resolution which ultimately leads to a higher uncertainty in the position of the object. In this way, they thought that such intermediate features could be used to improve the accuracy of CNNs without the penalties of speed and memory resources that were typical in other applications, as they are already part of the CNN pipeline and, hence, it is also independent of the backbone convolutional architecture. 
It is common that CNNs have several layers that output same feature size, such group of layers is known as a stage or block. The last layer feature map on a stage, denoted as $C_2,C_3,C_4,C_5$ for conv2, conv3, conv4 or conv5 outputs, have strides of 4, 8, 16, 32 pixels with respect to the input image. This last feature map, which corresponds on a Bottom-Up Pathway, is then used to create the so call Feature Pyramid Network (FPN). Each feature of the pyramid has:

A vertical connection (Top Down Pathway). As shown in Figure \ref{fig:FPN}, the feature maps at two levels are combined in order to merge the higher resolution of the early stage, i.e. more accurate location, with the stronger semantic information of the latter stage. As the latter stage feature map has a coarser resolution, an upsampling method (nearest neighbour for simplicity) is used to increase spatial resolution by a factor of 2. The ultimate objective of the top-down connections is to enrich the ealier stages with the stronger semantic information of the latter stages of the backbone convolutional model.

A lateral connection: Once that the vertical connection is computed, it is combined with the lateral connection, which merges feature maps of the same spatial size on an element-wise addition. The lateral connection input is generated after a 1x1 convolutional layer is applied to the corresponding Bottom-Up feature map. The aim for the 1x1 convolution is to reduce channel dimensions (d) to a fixed value of d=256. The ultimate objective of the lateral connection is to increase the location capabilities of the backbone model.
Thanks to the introduction of FPN, Lin et al. \cite{linFeaturePyramidNetworks2017}  improved Faster R-CNN AP@0.5 by 9.6 points by defining anchor areas of ${32^{2},64^{2},128^{2},256^{2}}$ and ${512^{2}}$ pixels on pyramid levels ${P_{3},P_{4},P_{5},P_{6},P_{7}}$ respectively and using anchor aspect ratios of ${1:2,1:1,2:1}$. Where $P_{i}$ represents a feature level with resolution $\frac{1}{2^{i}}$ of the input image. So if input resolution is 640x640, then $P_{3}^{in}$ represents feature level 3 with a resolution of 80x80 ($\frac{640}{2^{3}}=80$) and $P_{7}$ will have a resolution of 5x5 ($\frac{640}{2^{7}}=5$). Such inlet feature $P_{3}^{in}$ is combined with a top feature using an upsampling (factor x2) to create the final outlet feature prior to a convolution operation for feature processing as follows:

\begin{equation}
P_{3}^{out}=Conv(P_{3}^{in}+Resize(P_{4}^{out}))
\end{equation}

RetinaNet uses anchor areas of $\{32^{2},64^{2},128^{2},256^{2},512^{2}\}$ pixels on pyramid levels $\{P_{3},P_{4},P_{5},P_{6},P_{7}\}$  respectively and anchor aspect ratios of 1:0.5, 1:0.75, 1:1, 1:25 and 1:1.5, although this is a hyper-parameter that can be optimised if required. And for a denser scale coverage, anchor sizes of $\{2^{0},2^{\frac{1}{5}},2^{\frac{2}{5}},2^{\frac{3}{5}},2^{\frac{4}{5}}\}$ are added at each pyramid level (or octave), so in total 25 anchors per level are calculated on our analysis. In practice, $P_{2}$ is not used due to computational reasons.

\subsection{Focal Loss} \label{sec:focalloss}
Most of the Machine Learning (ML) problems are optimisation problems where we need either to maximise or minimise the output. NNs require a large number of labelled images where the algorithm iterates across the complete dataset and gives a prediction that it is compared against a Ground Truth (GT) label and an error is generated between the prediction and the actual value. This error is computed by a Loss Function (L) and minimized by an optimizer until an acceptable level is achieved. 

But as the loss function is non-convex, more than one local optima could exist, leading to poor model accuracy if the loss function topology is not explored conveniently. Hence, a careful selection of hyper-parameters is required to get good generalization capability and best performance on subsequent inference analysis. 

Hence, during this process, the NN will take a set of images or feature vectors ($x_{i}$) and map them to a resulting label, in case of a classification problem, by means of the dot product with the weight matrix W (omitting the bias term for brevity), resulting on our \emph{scoring function}:

\begin{equation}
s=f(x_{i},W)=Wx_{i}
\end{equation}

Traditionally, it was thought that two stage detectors, such as Faster R-CNN \cite{renFasterRCNNRealTime2016}, were slower but provided a higher accuracy. However, Lin et al. \cite{linFocalLossDense2018} investigated the reason of such behaviour and why Single Stage Detectors (SSD) performed worse. SSD detectors are very sensitive to the selection of anchor size and how such anchors cover the space of possible objects in the image. The conclusion of their investigation showed that SSDs are exposed to extreme class imbalance between foreground and background labels, where the latter case have much more examples than the earlier. In this situation, CNNs tend to be biased to predict the class with higher number of examples. This is not the case for R-CNN detectors, where the RPN reduces significantly the number of background samples but selecting a fixed number of samples, leading to a fixed foreground to background ratio of 1:3. On the other hand, SSDs need to evaluated a large number of candidate objects.

\begin{figure}
	\centering{}
	\includegraphics[scale=0.6]{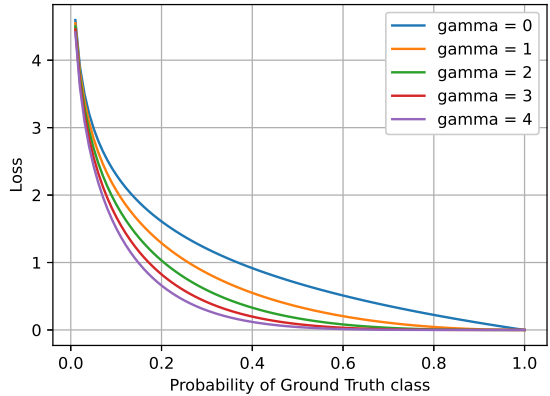}
	\caption{Focal Loss that adds a factor $(1-p_{t})^{\gamma}$ to the standard cross entropy criterion \cite{linFocalLossDense2018}}
	\label{fig:Focal}
\end{figure}

To compensate for such imbalance, Lin et al. \cite{linFocalLossDense2018} proposed a modification of the Cross Entropy Loss function, named as Focal Loss, by introducing a correction factor that hardly increases the loss function for the easy samples (i.e. down-weights inliers) but that penalises the more complex samples (i.e. up-weights outliers). See Figure \ref{fig:Focal}.

\begin{equation}
L=-\alpha_{t}(1-p_{t})^{\gamma}log(p_{t})
\end{equation}

where ${-\alpha_{t}log(p_{t})}$ is the weighted Cross Entropy (CE) loss and ${(1-p_{t})^{\gamma}}$ is a correction factor. For $\gamma=0$ the equation corresponds to the weighted CE (Figure \ref{fig:Focal} shows the graphical representation of the Focal Loss). The RetinaNet model is relatively robust and stable when $\gamma$ $\epsilon$ [0.5,5].
\subsection*{Optimizers\label{subsec:Optimizers}}
Optimizers are algorithms responsible for reducing the loss function in order to find its global minima and, hence, finding the most accurate results. They update the neural network weights and learning rate to achieve this goal. In our analysis, we have evaluated two optimizers: Stochastic Gradient Descent (SGD) and Adaptive Moment Estimation (ADAM) \cite{kingmaAdamMethodStochastic2017}.

SGD is a variant of the Gradient Descent algorithm that updates the model parameters more frequently, so instead of computing the whole dataset to define the way to go, it only takes a mini-batch or subset of the dataset. This results on high variance in model parameters and eventually can also overshoot the global minima. On the other hand, it accelerates the convergence process and requires less memory whilst it can also find new minima. The SGD can also be scaled based on a Learning Rate (LR) or $\alpha$, resulting in the following Weight matrix update expression on each iteration:

\begin{equation}
W=W-\alpha\nabla wf(w)
\end{equation}

Where $\alpha$ may change according to different strategies such as:

Step decay:
\begin{equation}
\alpha=\alpha_{0}\gamma^{floor(\frac{n}{s})}
\end{equation}

Time based:
\begin{equation}
\alpha=\frac{\alpha_{0}}{1+\gamma n}
\end{equation}

Exponential:
\begin{equation}
\alpha=\alpha_{0}\gamma^{n}
\end{equation}

Cosine:
\begin{equation}
\alpha=\frac{1}{2}\left(1+cos\left(\frac{n\pi}{N}\right)\right)\alpha_{0}
\end{equation}

At the beginning of training, all the NN weights are random values, or pre-trained on a different dataset, and hence far away from local minimum. Hence, applying a large learning rate may lead to instabilities. To avoid it, it is common to use a warm-up with reduced learning rate for the first epochs, as follows:

\begin{equation}
\alpha=\frac{n}{N}\alpha_{0}
\end{equation}

where n is the number of epochs and N is the total warm-up epochs.
Whatever the approach is, it is always recommended to start with a low learning rate, so the original model weights are not changed dramatically whilst still having enough capacity to get out of a local minimum. Once this warm-up epochs are exceeded, then the learning rate can be increased to look for the global local minimum. As mentioned above, at the same time, it is required to also start reducing its value as the convergence process continues to avoid overshooting the local minimum.

On the other hand, ADAM is an optimization of the SGD algorithm that combines several improvements: momentum acceleration and RMSPROP correction term. The great advantage of ADAM optimizer is its fast convergence rate and the ability of rectifying the vanishing learning rate and high variance produced during the iterative process, but at a cost of increase computational resources.

Momentum acceleration is a correction factor applied to SGD that increases the strength of the updates for those dimensions whose gradients point in the same direction whilst decrease the strength on those dimensions pointing in other directions. Momentum acceleration \cite{qianMomentumTermGradient1999} introduces a correction factor ($V_{t}$) to Cross Entropy equation, resulting on a faster convergence with smaller fluctuations on each epoch, and is defined as:

\begin{equation}
W=W-V_{t}=W-\gamma V_{t-1}+\alpha\nabla wf(w)
\end{equation}

Where the momentum term $\gamma$ is set usually to 0.9. But if the momentum term is too high, the algorithm may miss the local minima. So, additionally, a correction factor (another hyper-parameter in our model) can be applied: the Nesterov accelerated gradient. A more detailed explanation of Nesterov acceleration can be found in \cite{ruderOverviewGradientDescent2017}.

Momentum acceleration term:
\begin{equation}
v_{t}=\frac{\beta_{2}s_{t-1}+(1-\beta_{1})g_{t-1}}{1-\beta_{1}^{t}}
\end{equation}

RMSPROP term is responsible to take a moving average of squared gradients to keep an exponentially decaying average of past gradients that reduces the gradient velocity rate, according to the following expression:
\begin{equation}
s_{t}=\frac{\beta_{1}v_{0-1}+(1-\beta_{2})g{}_{t-1}^{2}}{1-\beta_{2}^{t}}
\end{equation}

ADAM Gradient expression:
\begin{equation}
w_{t}=w_{t-1}-\alpha\frac{v_{t}}{\sqrt{s_{t}}+\varepsilon}
\end{equation}

with typical values defined for bias correction factors are $\beta_{1}=0.9$, $\beta_{2}=0.999$ and a residual value of $\varepsilon=10^{-8}$ is used to prevent division by zero.

\subsection{Performance metrics} \label{subsec:Detection-Performance-metrics}
\begin{figure}
	\centering{}
	\includegraphics[scale=0.2]{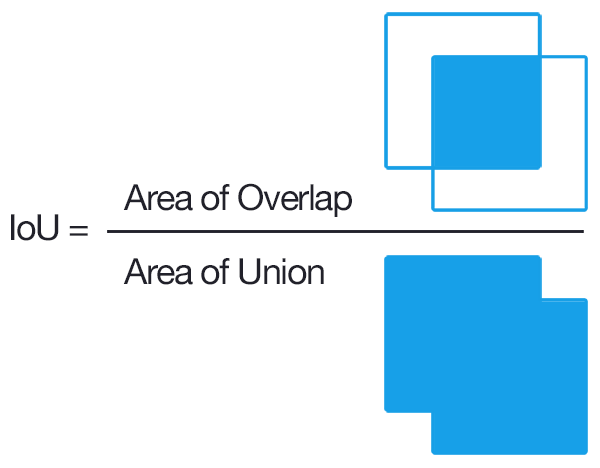}\caption{Definition of IoU}
	\label{fig:IoU}
\end{figure}

While the NN is being trained or once the global minima was found, it is required to measure the accuracy of the model. As mentioned in the introduction, object detection involves defining bounding boxes around the detected objects. Hence, a dedicated metric called Interception over Union (IoU), also known as the Jaccard Index, is used to measure the level of over-lapping between the model inference Bounding Box (BBox) and the Ground Truth (GT) as shown in Figure \ref{fig:IoU}. A typical value or threshold for IoU is 0.5, so if the model IoU is higher than this threshold, then it is considered that it has properly located an object.

\begin{equation}
IoU=\frac{OverlapArea}{UnionArea}\label{eq:IoU}
\end{equation}

In addition to IoU, it is also important to mention the \textbf{Confidence Score} or the probability that a bounding box contains an object. This value is the output of the classifier part of the NN. As a hyper-parameter, there is an associated confidence threshold, so objects whose confidence score is above such threshold will be accepted as a match to the Ground Truth.

\textbf{Average Precision (AP)} in case of an object detection problem is defined as the number of Positives divided by the number of queries to the model. Hence, \textbf{Precision} or Positive Predictive Value (PPV) can be defined as \emph{the ability to identify only the relevant objects} and it is defined as the ratio between the True Positives (TP) and all the positive predictions from the model and is defined as:

\begin{equation}
Precision=\frac{TP}{TP+FP}\label{eq:precision}
\end{equation}

\textbf{Recall} or Sensitivity or True Positive Rate (TPR) values will corresponds to number of positives evaluated so far at evaluation is divided by the total number of positives, i.e. \emph{the ability of the model to find all relevant objects:}

\begin{equation}
TPR=Recall=Sensitivity=\frac{TP}{TP+FN}\label{eq:recall}
\end{equation}

\textbf{Mean Average Precision (mAP)} is simply the average of all AP after N interrogations to the model and provides a simple metric to compare different models, and it is defined as:

\begin{equation}
mAP=\frac{1}{N}\sum_{i=1}^{N}AP_{i}
\end{equation}

The above definition is right when measuring the performance of a classification model. However, when dealing with object detection model we need to reconsider what is a True Positive (TP), a False Positive (FP) or a False Negative (FN). The True Negative (TN) is not evaluated in this case as it is always assume that each image will have a background area on it. Hence, the following cases will appear:

\begin{itemize}
\item TP will be considered only when two conditions are met: BBox ${IoU>0.5}$ (or whatever threshold is defined for the calculation) with regards the GT BBox AND the Confidence Score is higher than the threshold.
\item FP will be considered under two potential scenarios: ${IoU<0.5}$ (or whatever threshold is defined for the calculation) even if the Confidence Score was higher than the Confidence Threshold OR there is duplicated BBox.
\item FN will be considered also under two potential scenarios: No detection	at all of the object (${IoU<threshold}$) OR if the object is classified with the wrong label (or Confidence Score ${<}$ Confidence Threshold) even if ${IoU>0.5}$.
\end{itemize}

%
%
%
\begin{figure*}
	\includegraphics[width=0.50\textwidth]{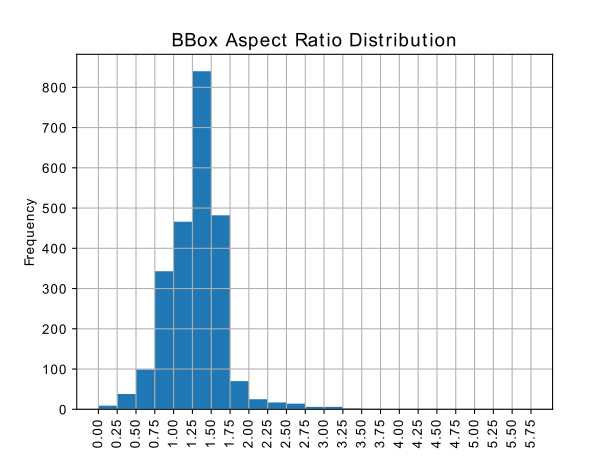}
	\includegraphics[width=0.50\textwidth]{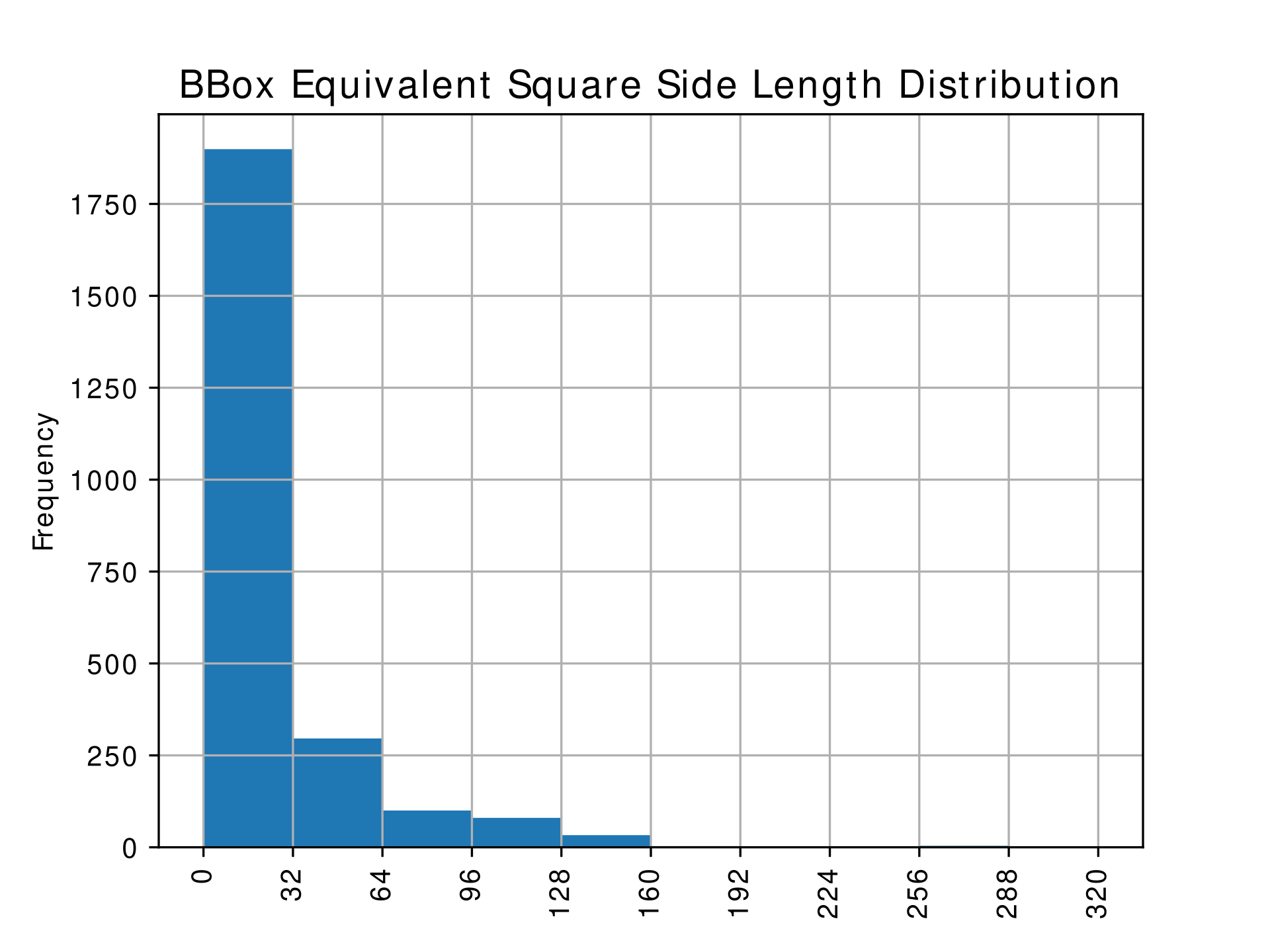}
	\caption{GDXray dataset analysis}
	\label{fig:GDXray-dataset-key}       
\end{figure*}

\begin{figure}
	\centering{}%
	\includegraphics[width=1.0\textwidth]{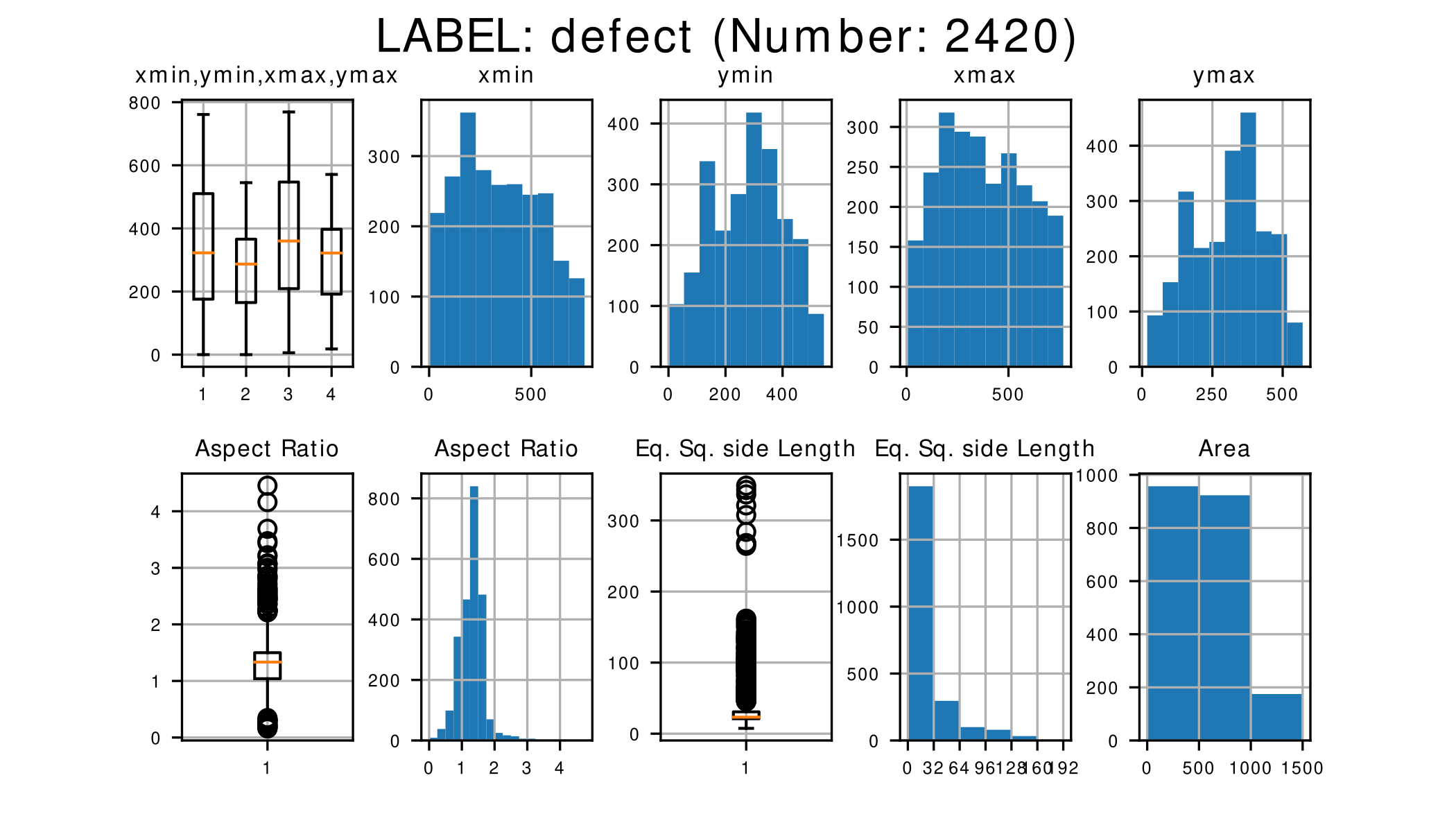}
	\caption{GDXray defect location and outliers sizes
		\label{fig:GDXray-defect-location}}
	
\end{figure}

\section{Experimental results}
\subsection{Dataset analysis}
GDXray Castings dataset \cite{meryGDXrayDatabaseXray2015a} contains 2727 X-ray images mainly from automotive aluminium parts, although around 2420 images are properly labelled with BBox defining the exact position of the defect. Most of the images were taken by BAM Federal Institute for Materials Research and Testing (Berlin, Germany). The original 12 bit images were scanned using a LASER scanner LS85 SDR with a pixel size of 40.3 micros (630 dpi) and were transformed to 8 bits using a linear Look-up-Table (LUT) proportional to the optical film density. All images where converted into JPG RGB format with varying sizes: 350x151, 768x572, etc.

As can be seen in Figure \ref{fig:GDXray-dataset-key}, most of the defects in this dataset can be considered as small (less than 32 pixels) where the defect aspect ratio of the Ground Truth boxes is mainly between 1.0 and 1.75. Flaws are located all over the images as deduced from Figure \ref{fig:GDXray-defect-location}, that also shows a boxplot with the defect outliers in terms of size and aspect ratio. All this information was used to optimise the RetinaNet model to this particular dataset. The following subsections will show the result of running dozens of experiments where an orthogonal hyper-parameters optimisation approach was followed till the best accuracy combination was finally found. Table \ref{tab:anchor_size} shows the initial baseline analysis (experiment 1) and the influence of anchor size on model accuracy where a significant improvement can be observed for anchor size 2 (experiment 2, run with 640x640 input image size, 25\% test size, SGD optimizer, cosine decay learning rate and horizontal flip as data augmentation).

\begin{table}
	\caption{Influence of anchor size selection}
	\label{tab:anchor_size}
	\centering{}
	\begin{tabular}{ccc}
		\hline\noalign{\smallskip} 
		Exp & Anchor Size & mAP@IoU=50\%\\
		\noalign{\smallskip}\hline\noalign{\smallskip}
		1 & 1.0 & 0.759\\
		2 & 2.0 & 0.903\\
		3 & 3.0 & 0.898\\
		\noalign{\smallskip}\hline
	\end{tabular}%
\end{table}

\subsection{Influence of dataset size}
NNs require a large number of images to find the weights that lead to the loss function global minima, even if the model is already pretrained on another dataset. In our case, we started the adaptation of the RetinaNet model to the GDXray dataset using fine tuning from a pretrained model based on Microsoft COCO dataset\footnote{See https://cocodataset.org/\#home}, that contains 2.5 million labelled instances in 328k RGB images of 91 object types that would be easily recognizable by a 4 year old child \cite{linMicrosoftCOCOCommon2015}.

\begin{figure}
	\centering{}%
	\includegraphics[scale=0.50]{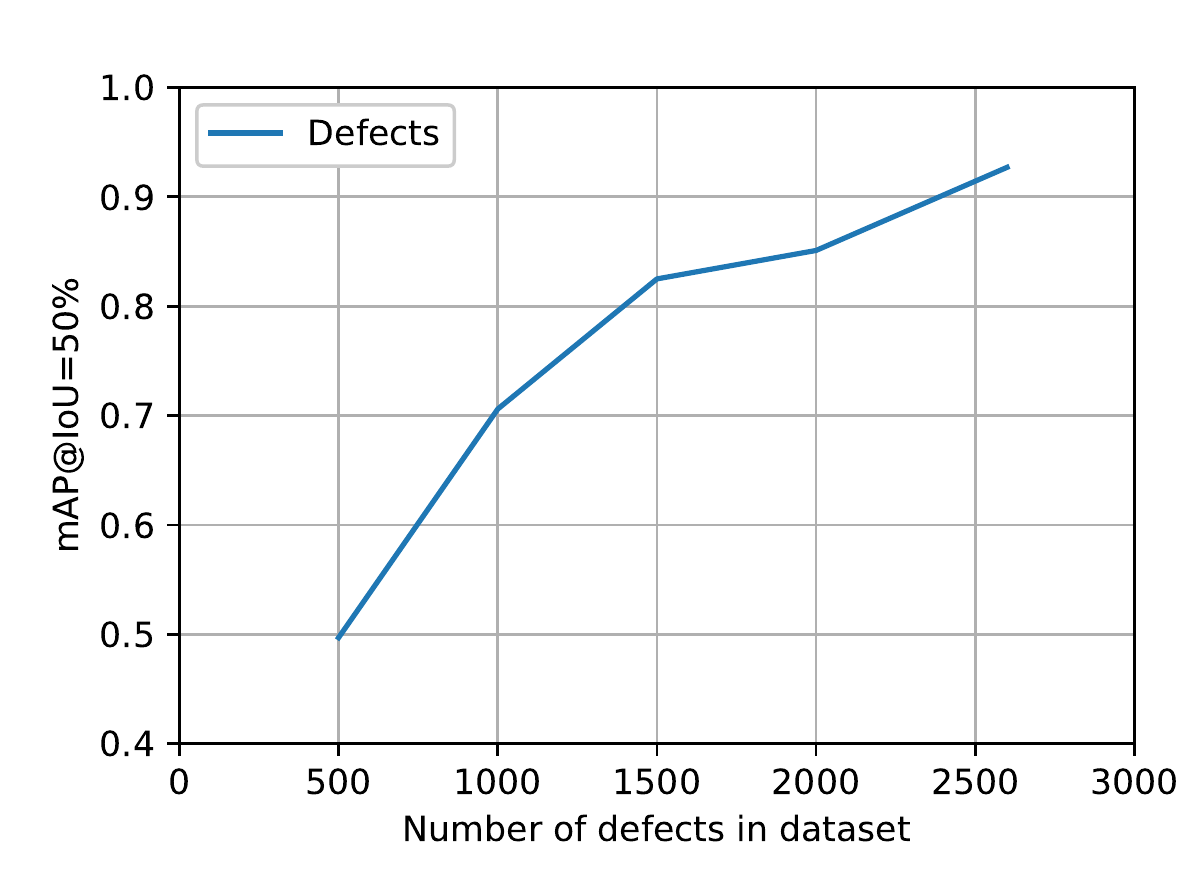}
	\caption{Influence of dataset size (number of defects) in achieved accuracy}
	\label{fig:number-images}       
\end{figure}

Figure \ref{fig:number-images} shows how the accuracy of the RetinaNet model improves as more semantic information is added into the training. At a constant test dataset size of 25\%, a logarithm trend is observed, so there is a critical dataset size where good accuracy is achieved with a reasonable number of images. Above this size, a large increment of defects will be required to achieve a residual accuracy increase in the model. This is an important conclusion as it will define the optimum size in industrial production applications where the cost of acquiring and labelling new images is usually very high in terms of available resources and time schedule. 

Hence, as a general rule, a few thousand images (defects) are required for each class in order to achieved acceptable accuracy results. In the particular case of the GDXray dataset, composed by single label defects, it suggests that it could be optimised by adding a few hundreds of defects and this will allow to increase future models accuracy.

\subsection{Influence of dataset split}
When dealing with NNs, it is very important to explore how to split the dataset to optimise the model. Traditionally, the dataset is divided into three subsets: training, validation and test. Training dataset will be used to train all the NN weights by minimizing the loss function and employing an optimizer to reach, ideally, the global minimum of such function.  The Validation dataset needs to be independent of the training and will be used during the optimization process to evaluate the metrics and convergence criteria. Last, but not least, the test dataset, must be independent of the training/validation datasets and will be used to demonstrate the generalization capability of the trained model.

An important hyper-parameter is how the data is split between training and test datasets. Table \ref{tab:Influence-of-training} shows the achieved model accuracy depending on the percentage of defects that was allocated to the test dataset, with all the rest of the hyper-parameters being the same. 

Simultaneously, it was also evaluated the effect of the input image size to the RetinaNet model. NNs models require to introduce square images in order to optimise the algebraic calculations behind the optimizer algorithm. In practice, that means that a resizing of the original images is required to fit the target input size. The higher such size is, the higher the computational cost of the training will be. In our evaluation, we have explore two model input sizes: 640x640 and 1024x1024 pixels.

Table \ref{tab:Influence-of-training} shows that at least 2.5\% accuracy improvement is achieved by increasing the model input size up to 1024x1024 pixels. As most of the GDXray dataset image size is in the range of 350x151 and 768x572 pixels, we think that upsizing the images permits to enlarge the smallest defects and, hence, making them more visible to our model, resulting in improved accuracy.

On the other hand, for this particular dataset, it seems that using 25\% of the available defects for testing and 75\% for model training is around the optimum performance.

Increasing training size will add more features to learn to the model but, on the other hand, will be available to generalize worse if the additional defects are not providing extra variance information. However, it test size is excessive, the effective number of images to be used by the model will be equivalent to moving to the left on Figure \ref{fig:number-images} where the slope is higher, resulting on lower accuracy results.

\begin{table}
	\caption{Influence of test dataset size. Left (640x640 input image), Right(1024x1024 input image)}
	\label{tab:Influence-of-training}
	\begin{centering}
		\begin{tabular}{ccc}
			\hline\noalign{\smallskip} 
			Exp & Test Size & mAP@IoU=50\%\\
			\noalign{\smallskip}\hline\noalign{\smallskip}
			2 & 25\% & 0.903\\
			30 & 40\% & 0.896\\
			31 & 15\% & 0.886\\
			\noalign{\smallskip}\hline
		\end{tabular}%
		\begin{tabular}{ccc}
			\hline\noalign{\smallskip}
			Exp & Test Size & mAP@IoU=50\%\\
			\noalign{\smallskip}\hline\noalign{\smallskip}
			11 & 25\% & 0.927\\
			33 & 40\% & 0.922\\
			32 & 15\% & 0.915\\
			\noalign{\smallskip}\hline
		\end{tabular}
		\par\end{centering}
\end{table}

\subsection{Influence of regularization techniques}
A typical problem that appear during training ML models is over-fitting, where a model accuracy is quite high on training but when inference is applied to a similar dataset, a very poor accuracy is achieved. This is the consequence of the model learning or capture the dataset features but also its noise, e.g. features that do not provide any information and that are caused by randomness. Hence, in this type of NN models, it is quite common to use regularization techniques defined as “any modification we make to a learning algorithm that is intended to reduce its generalization error, but not its training error”, according to Goodfellow et al.\cite{goodfellowGenerativeAdversarialNetworks2014}. 

Data augmentation is a very common regularization technique that involves generating new training samples from an original image by increasing the number of input images. Such new images are obtained by applying cropping, rotation, mirroring, changes in scale, shearing, colour shifts etc. By introducing such changes in the input image, we also introduce some random jitter points in a normal distribution, making the NN more robust and increasing its capability to generalize and, hence, reducing the risk of over-fitting the input data.

Data augmentation is a traditional technique employed for data regularization that helps the CNN to generalize better based on a reduced set of images. Several options were explored as shown in Table \ref{tab:augmentation}. First approach was to use random horizontal flipping (HFlip) of the images, that leads to 0.914 mAP. However, adding vertical flipping to this case (HVFlip), which consists on putting the image upside-down, actually penalises the accuracy as it is seldom likely to appear on casting defects. Adding crop compensated the penalising effect of vertical flip but restored the accuracy to the initial level. When random scale of the images was added, as significant penalization was introduced, reducing accuracy up to 0.889 mAP. Hence, it is concluded that adding too much data augmentation or inappropriate techniques may penalise results. A case by case evaluation is required for other datasets.

\begin{table}
	\caption{Influence of data augmentation}
	\label{tab:augmentation}
	\centering{}
	\begin{tabular}{ccc}
		\hline\noalign{\smallskip} 
		Exp & Augmentation & mAP@IoU=50\%\\
		\noalign{\smallskip}\hline\noalign{\smallskip}
		22 & HFlip & 0.914\\
		23 & HVFlip & 0.906\\
		24 & HVFlip + Crop & 0.911\\
		25 & HVFlip + Crop+ scale& 0.889\\
		\noalign{\smallskip}\hline
	\end{tabular}%
\end{table}

\subsection{Influence of optimizer and loss function}
During the CNN, it is convenient to start with a large learning rate ($\alpha$) in order to approximate faster to the (ideally) global minimum of the loss function, whilst when approaching the minimum, a reduction on the learning rate is required to avoid oscillations around this point. In all our experiments, we used 2500 iterations as warm-up to reduce initial change of weights and a cosine decay learning rate.

Table \ref{tab:optimizer} shows the influence of the optimizer selection on model accuracy. Experiment 6 is the baseline using SGD, whilst experiment 7 is just changing the optimizer to ADAM. In this case, it is observed an accuracy increase of 5\% mAP. However, this data is significantly reduced on experiments 20 and 22, where the focal loss gamma was increased up to a value of 4. However, ADAM still shows a positive improvement on model accuracy. 

Table \ref{tab:optimizer} shows the influence of $\gamma$ factor on focal loss. It can be observed that making the training harder, by increasing the $\gamma$ factor, leads to a model 3\% accuracy improvement for experiment 9. However, such improvement is vanished in the more accurate configuration corresponding to experiments 21 and 22 where ADAM optimizer was used.

\begin{table}
	\centering{}%
	\caption{Influence of optimizer}
	\label{tab:optimizer}
	\begin{tabular}{ccc}
		\hline\noalign{\smallskip}
		Exp & hyper-parameter & mAP@IoU=50\%\\
		\noalign{\smallskip}\hline\noalign{\smallskip}
		6 & Baseline SGD & 0.75\\
		7 & As Exp. 6 + ADAM & 0.803\\
		20 & Baseline SGD + focal loss $\gamma=4$ & 0.907\\
		22 & as Exp. 20 + ADAM & 0.915\\
		\noalign{\smallskip}\hline
	\end{tabular}
\end{table}

\begin{table}
	\centering{}%
	\caption{Influence of focal loss $\gamma$}
	\label{tab:loss}
	\begin{tabular}{ccc}
		\hline\noalign{\smallskip}
		Exp & hyper-parameter & mAP@IoU=50\%\\
		\noalign{\smallskip}\hline\noalign{\smallskip}
		8 & Baseline SGD+ focal loss $\gamma=2$ & 0.752\\
		9 & As Exp. 8+ focal loss $\gamma=4$ & 0.785\\
		21 & Baseline + focal loss $\gamma=2$ & 0.914\\
		22 & as Exp. 21 + focal loss $\gamma=4$ & 0.915\\
		\noalign{\smallskip}\hline
	\end{tabular}
\end{table}

\subsection{Influence of number of GPUs}
Table \ref{tab:Influence-of-number} investigates the influence of the number of GPUs in the final accuracy and total training time. All the experiments shown in this papers were run on a computer with 16 CPUs Inter(R) Xeon(R) E5-2640 v4 @ 2.10 GHx and on NVIDIA GTX1080Ti GPUs with 11 GB RAM memory.

\begin{table}
	\centering{}%
	\caption{Influence of number of GPUs involved during training}
	\label{tab:Influence-of-number}
	\begin{tabular}{cccc}
		\hline\noalign{\smallskip}
		Exp & hyper-parameter & Time@40000 steps & mAP@IoU=50\%\\
		\noalign{\smallskip}\hline\noalign{\smallskip}
		40 & 1 GPU (batch\_size=2) & 452 min & 0.921\\
		41 & 2 GPU (batch\_size=2) & 287 min & 0.876\\
		42 & 3 GPU (batch\_size=6) & 548 min & \textbf{0.942}\\
		\noalign{\smallskip}\hline
	\end{tabular}
\end{table}

As the number of GPUs is increased, the batch size can be increased, leading to a smoother total loss evolution, with less peaks and valleys. This is caused by the fact that more images are used in each step, leading to a more homogeneous distribution across different batches, leading to a smoother curve. If the batch size is keep constant, duplicating the number of GPUs (experiment 41 vs experiment 40) leads to a significant time reduction, although it is never achieved a x2 factor. This is because by adding more GPUs there is also more overhead added in the process associated to the communication amongst the different GPUs. This can lead to a bottleneck, where the GPUs computations are completed and the GPUs is waiting at idle while it receives a new set of data. 

\begin{figure}
	\centering{}
	\begin{tabular}{cc}
		\includegraphics[scale=0.10]{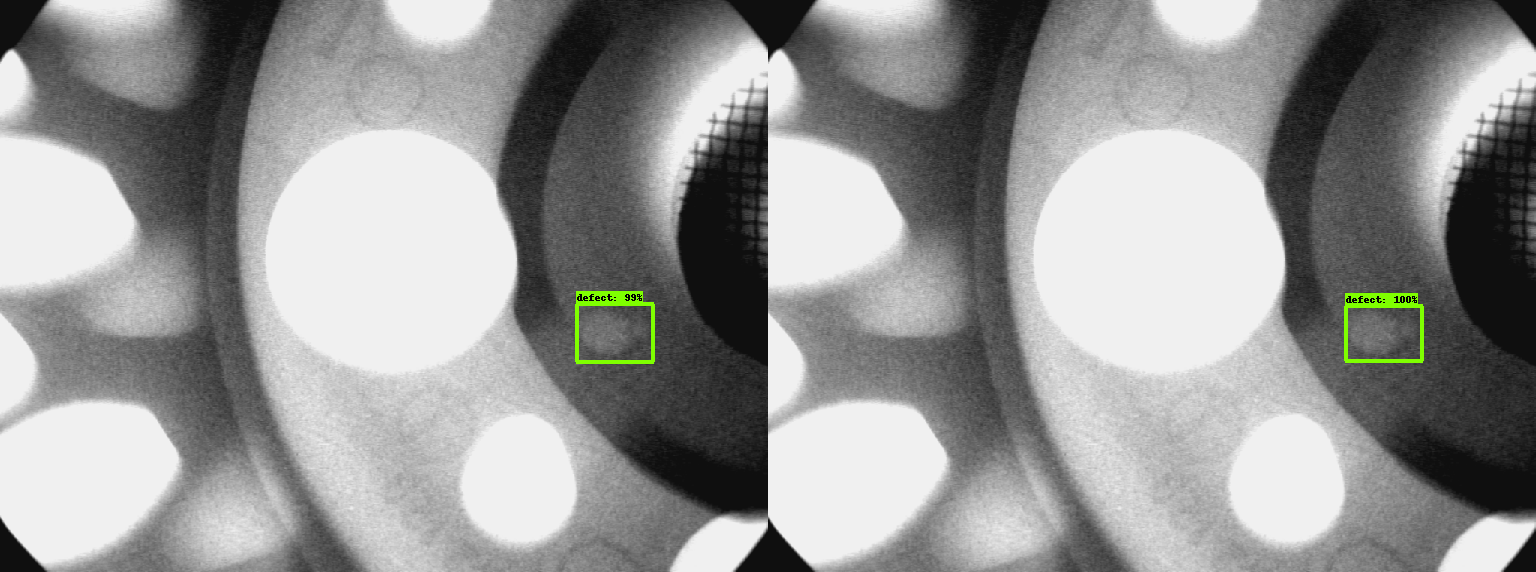} & \includegraphics[scale=0.10]{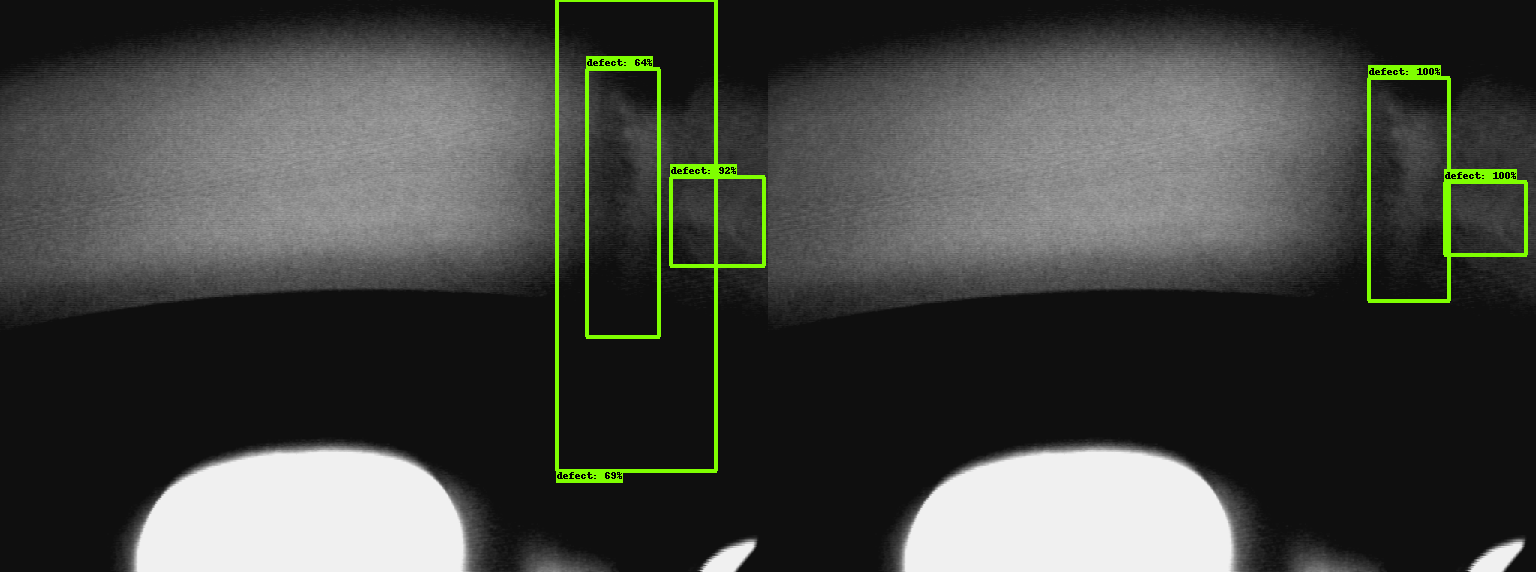}\\
		\includegraphics[scale=0.3]{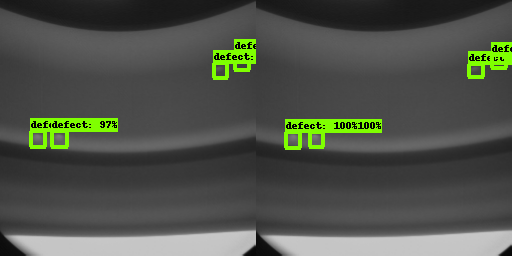}  & \includegraphics[scale=0.3]{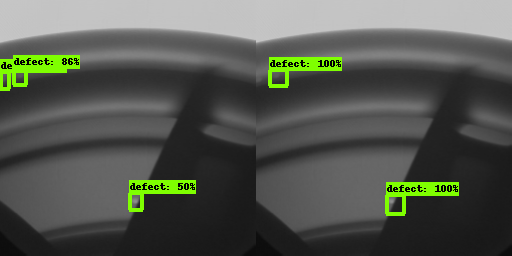}\\
		\includegraphics[scale=0.10]{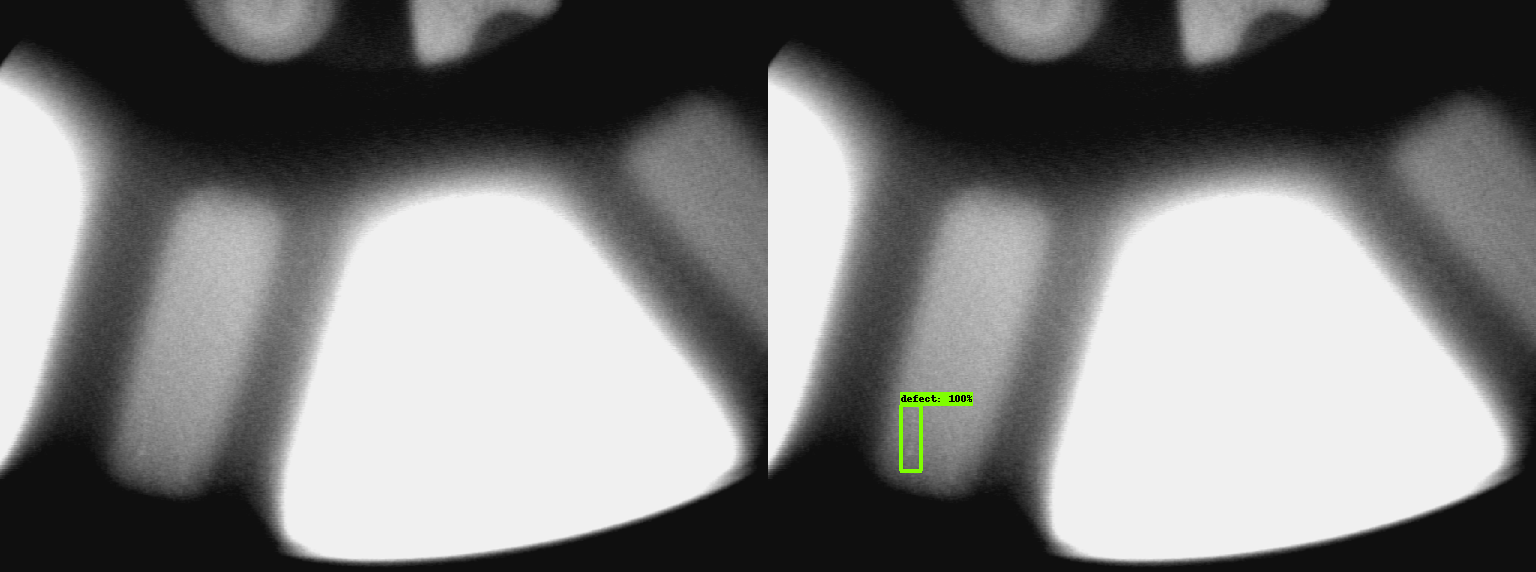} & \includegraphics[scale=0.10]{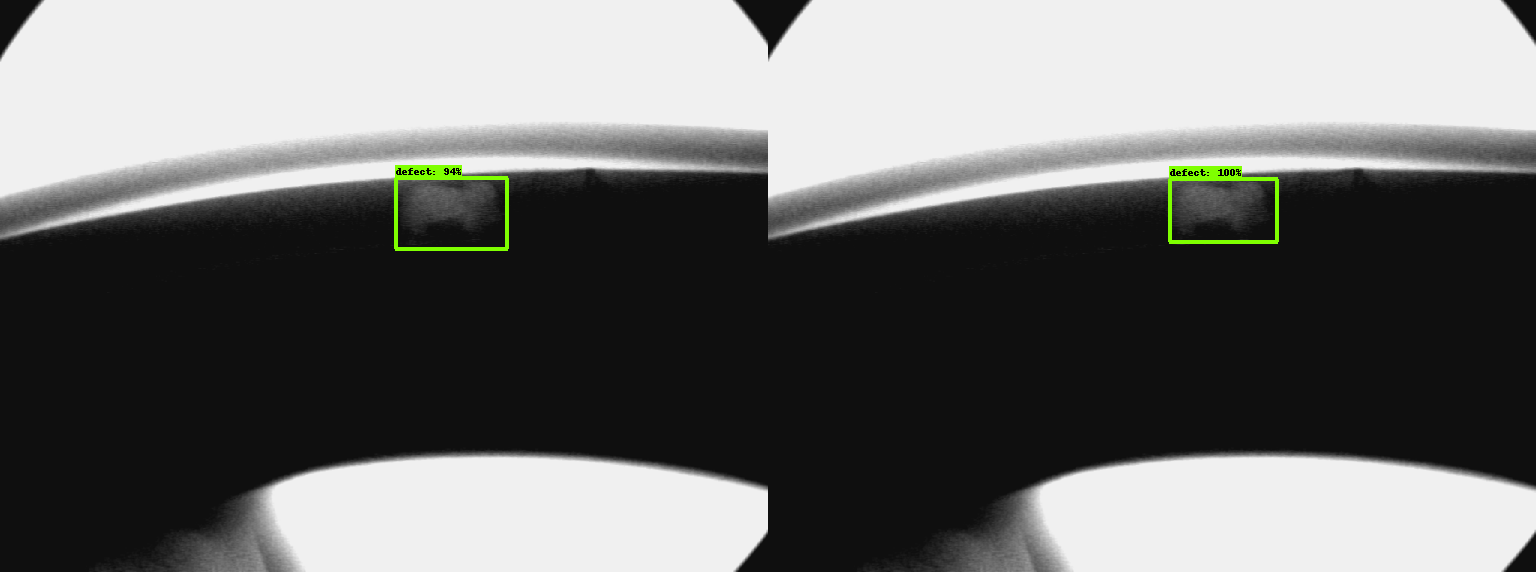}\\
		\includegraphics[scale=0.10]{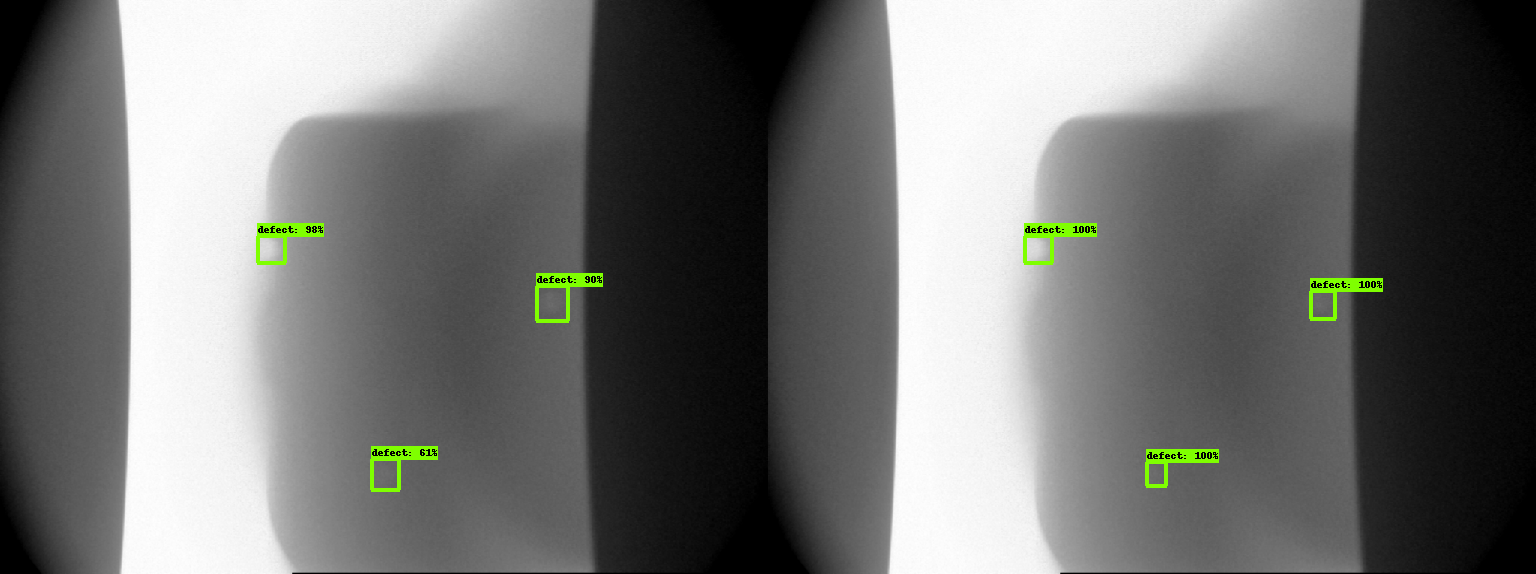} & \includegraphics[scale=0.10]{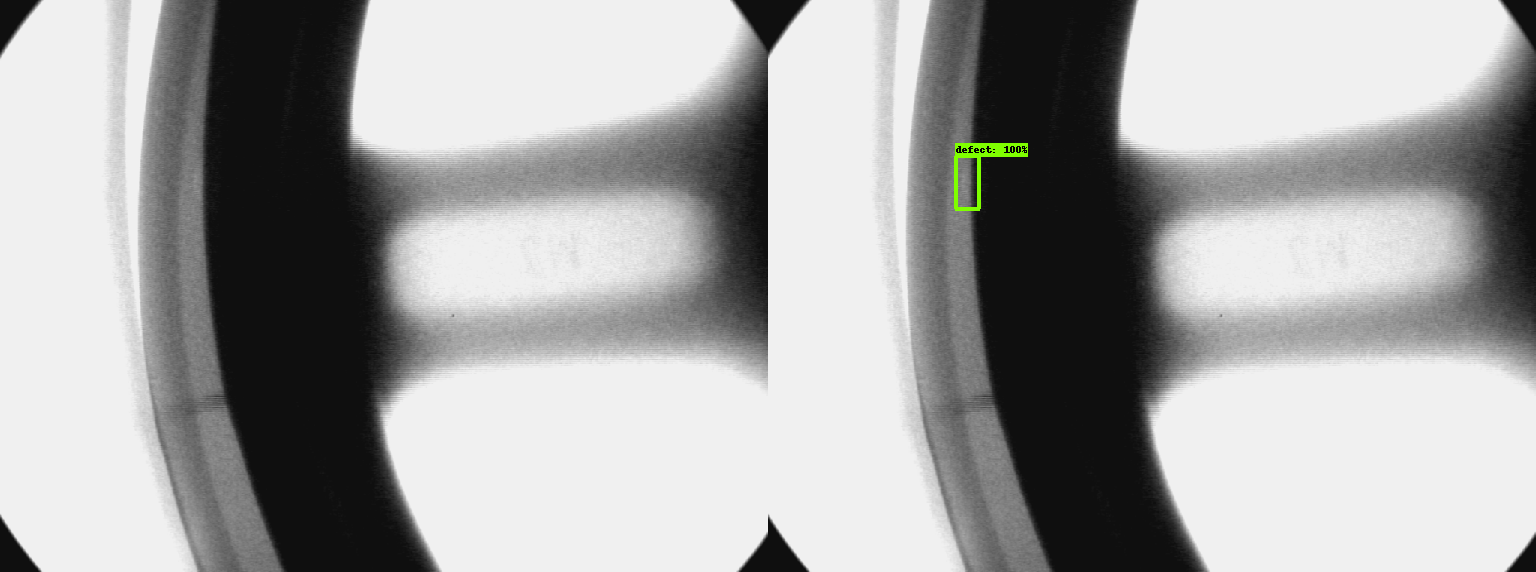}\\
		\includegraphics[scale=0.10]{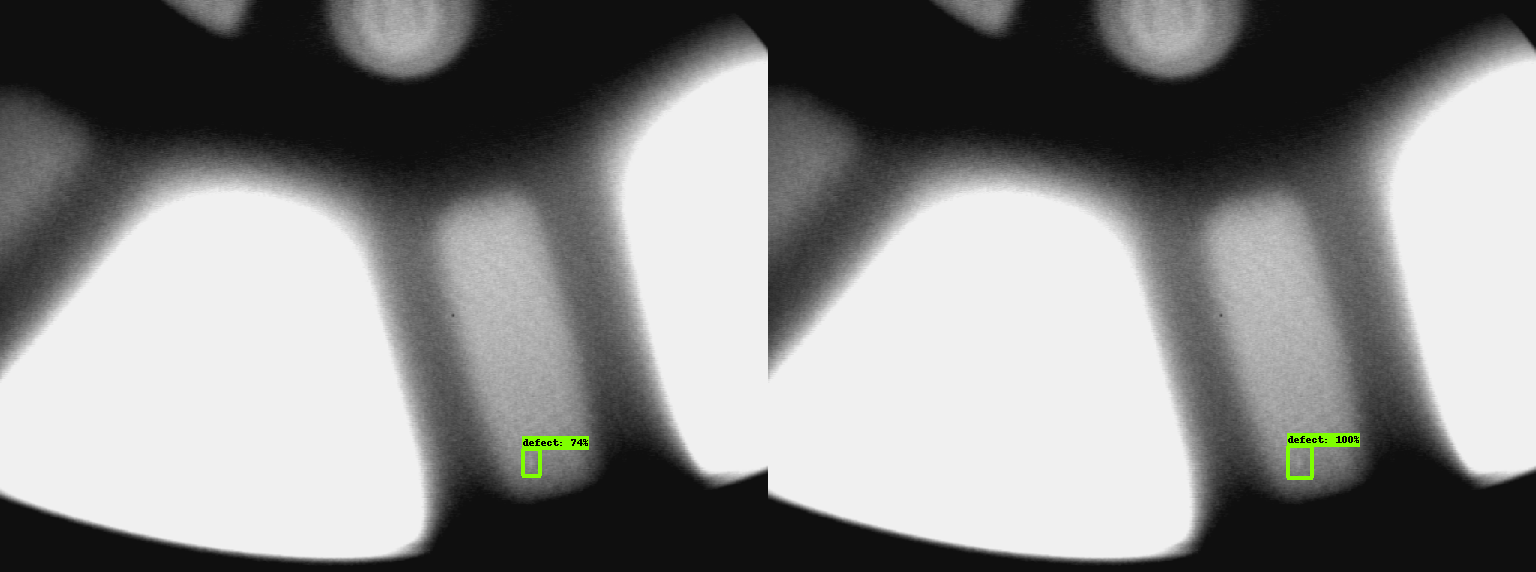} & \includegraphics[scale=0.10]{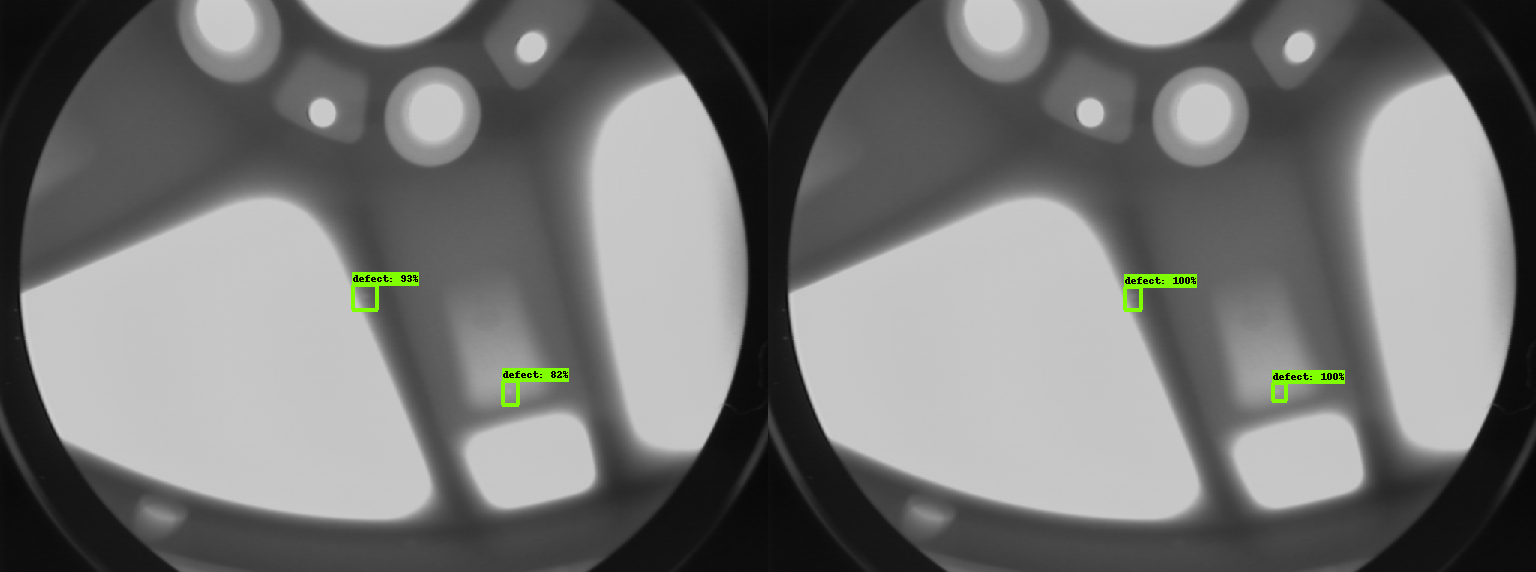}\\
		\includegraphics[scale=0.18]{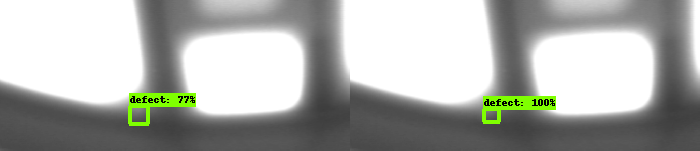} & \includegraphics[scale=0.18]{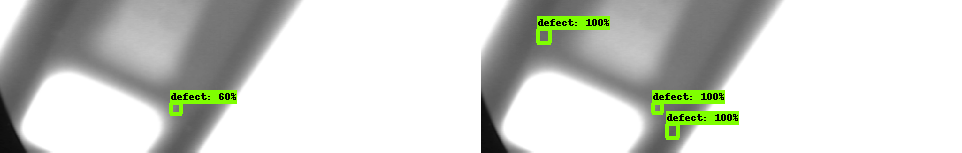}\\
		\includegraphics[scale=0.10]{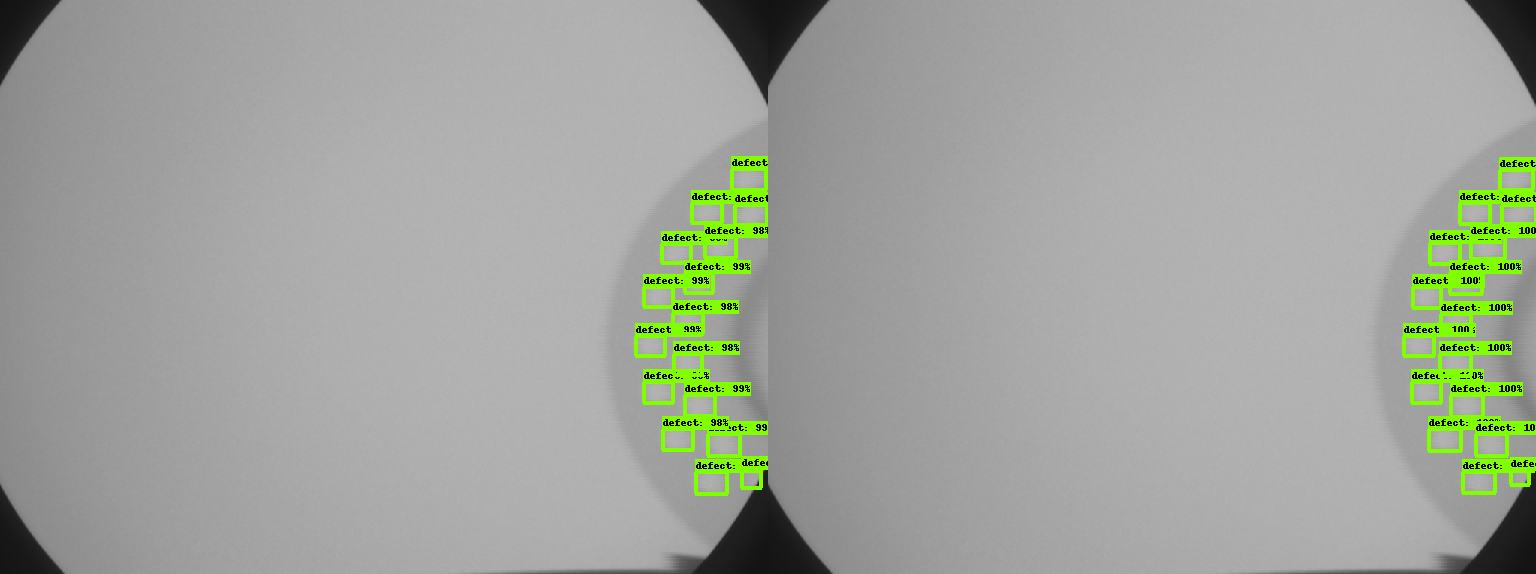} & \includegraphics[scale=0.10]{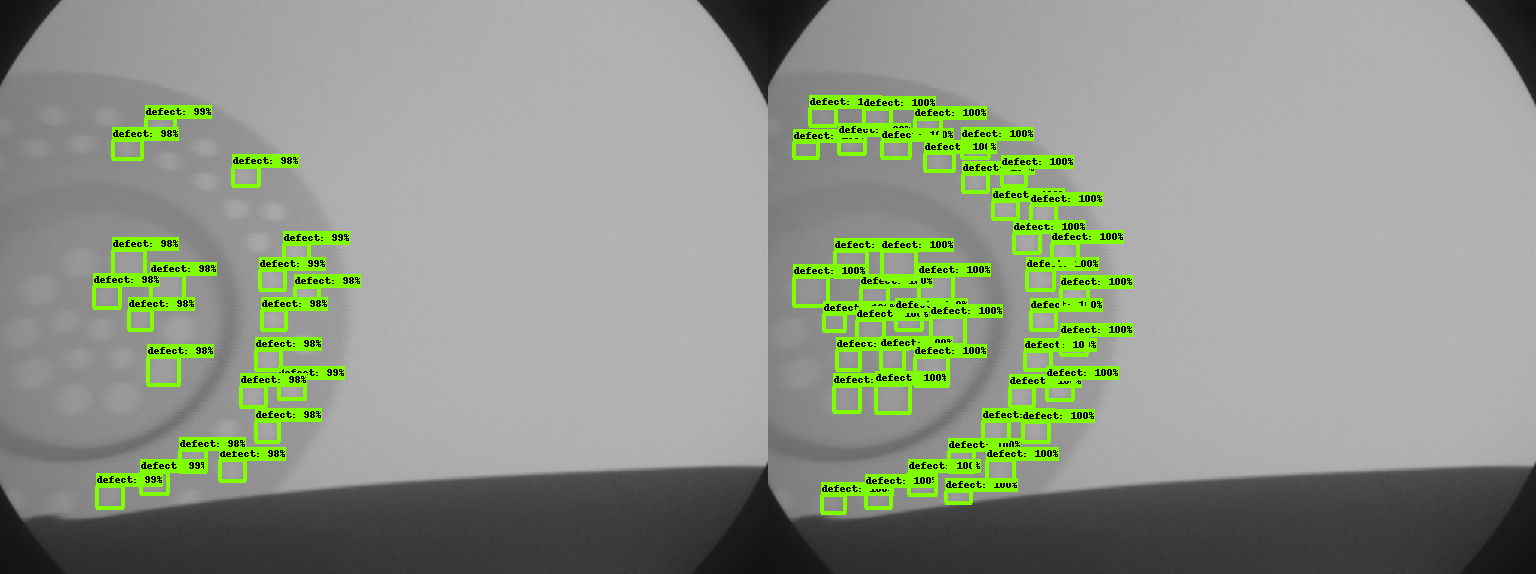}\\
	\end{tabular}
	\caption{Samples of achieved accuracy (I): Model results (left) vs Human inspector (right)}
	\label{fig:inference1}
\end{figure}

\begin{figure}
	\centering{}
	\begin{tabular}{cc}
		\includegraphics[scale=0.10]{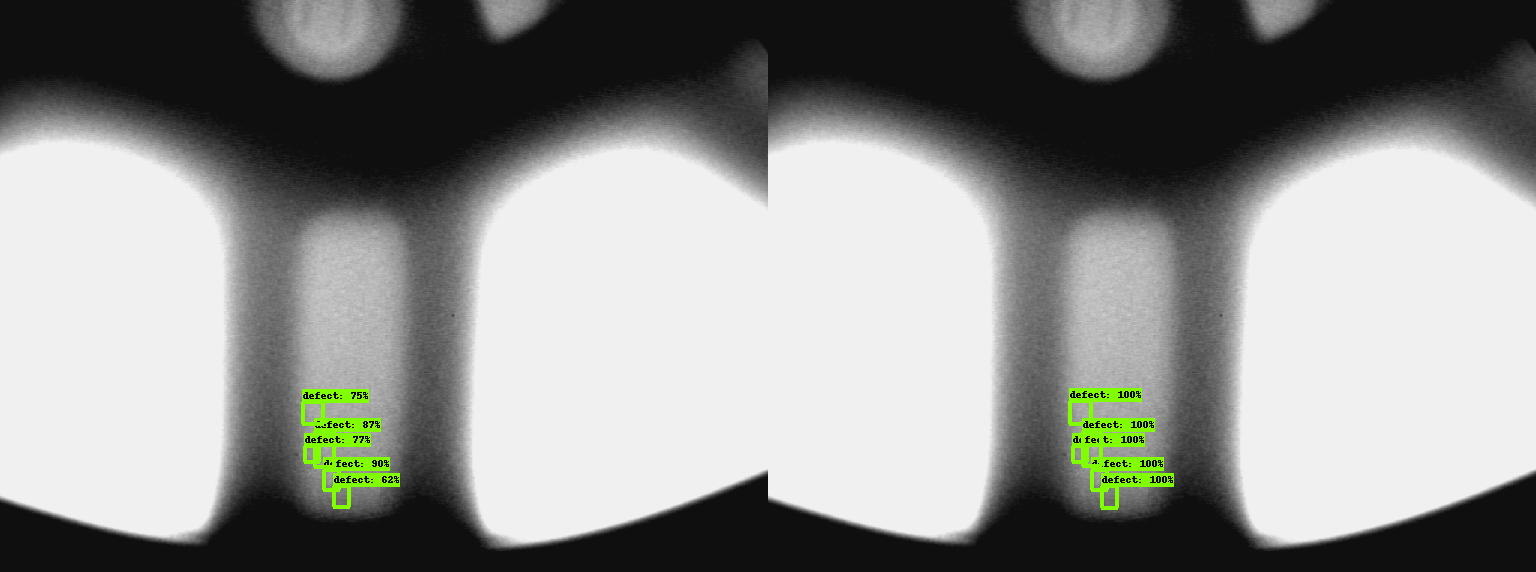} & \includegraphics[scale=0.10]{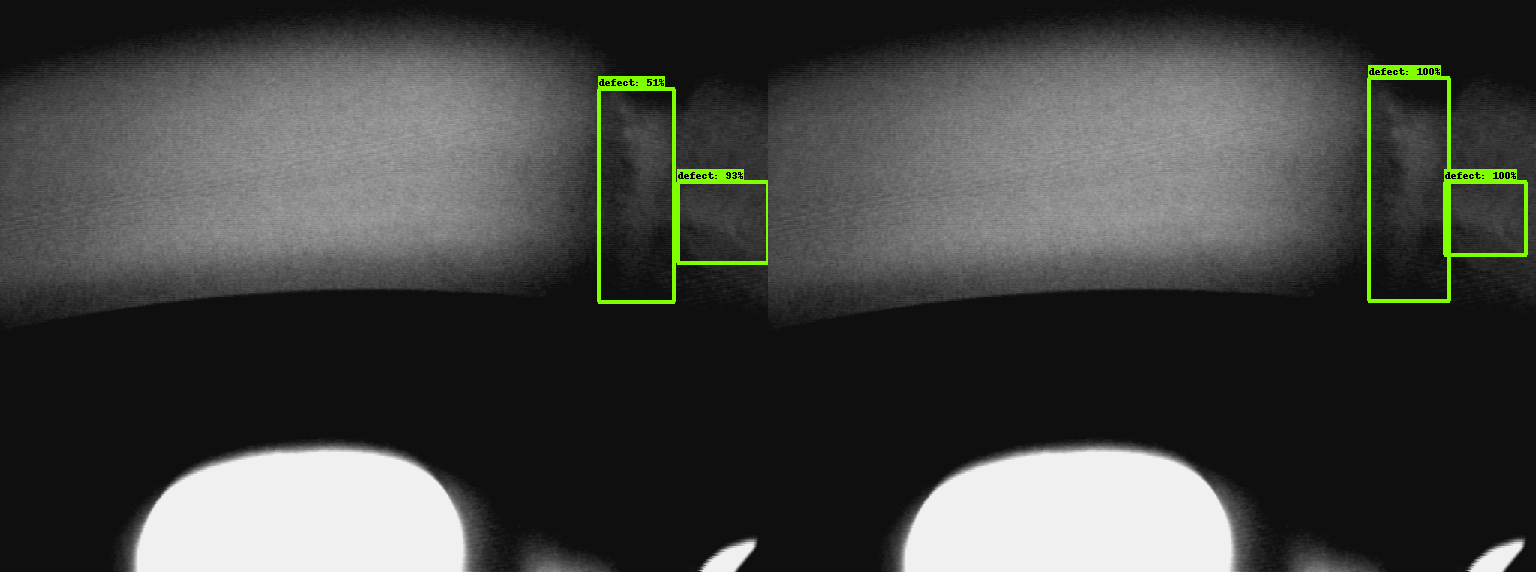}\\
		\includegraphics[scale=0.3]{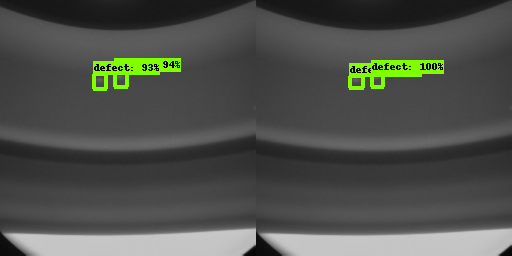}  & \includegraphics[scale=0.10]{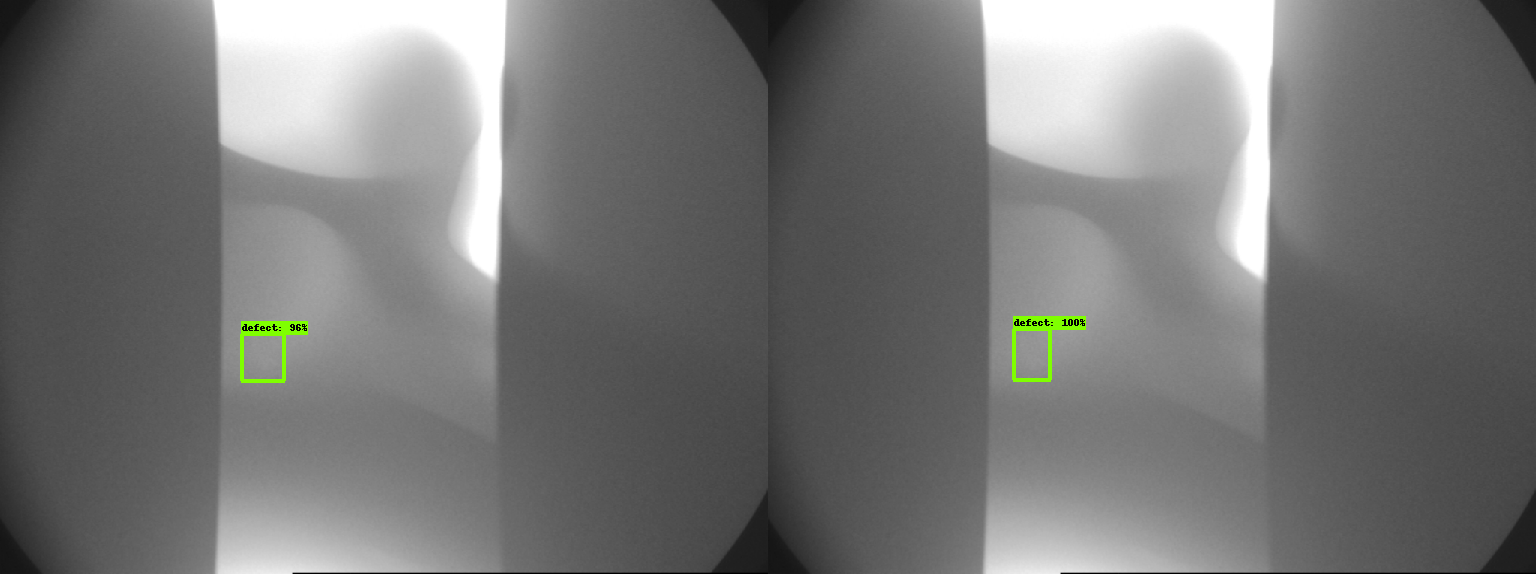}\\
		\includegraphics[scale=0.10]{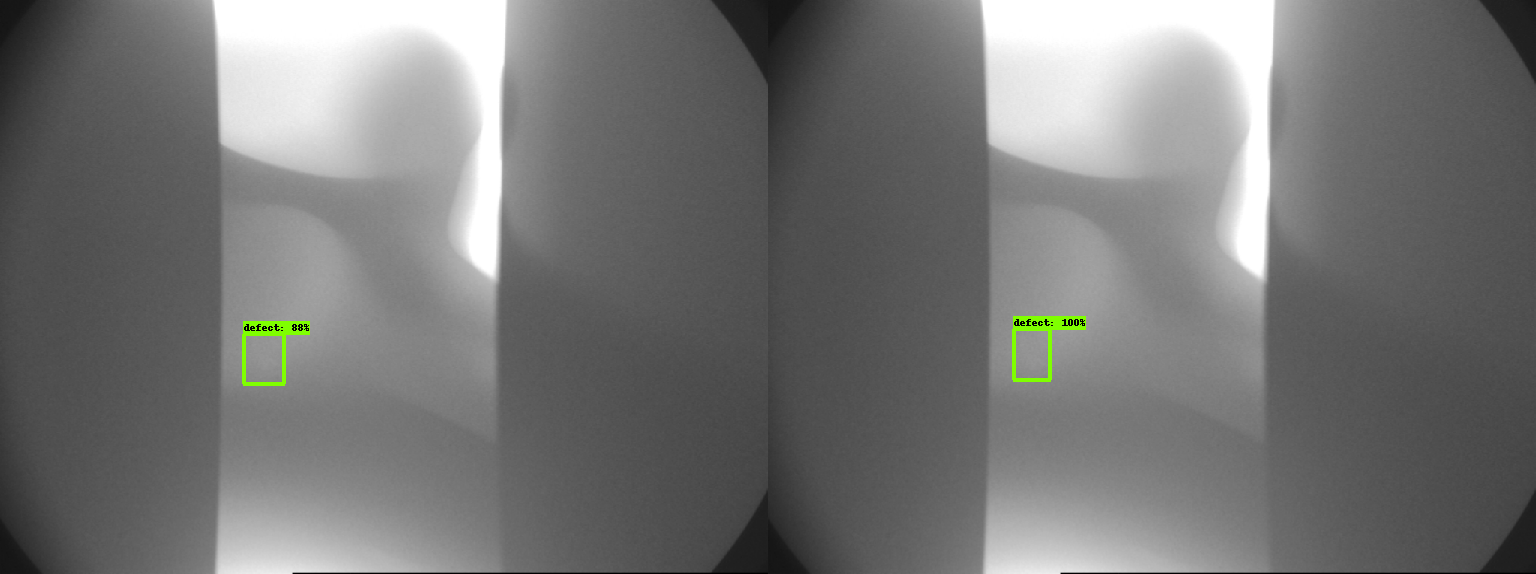} & \includegraphics[scale=0.30]{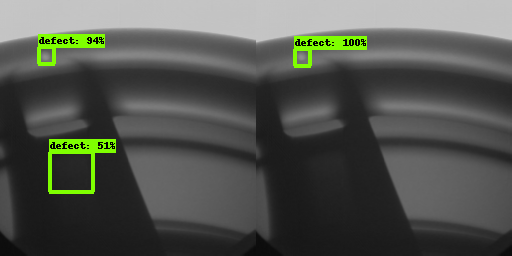}\\
		\includegraphics[scale=0.10]{imageData12} & \includegraphics[scale=0.10]{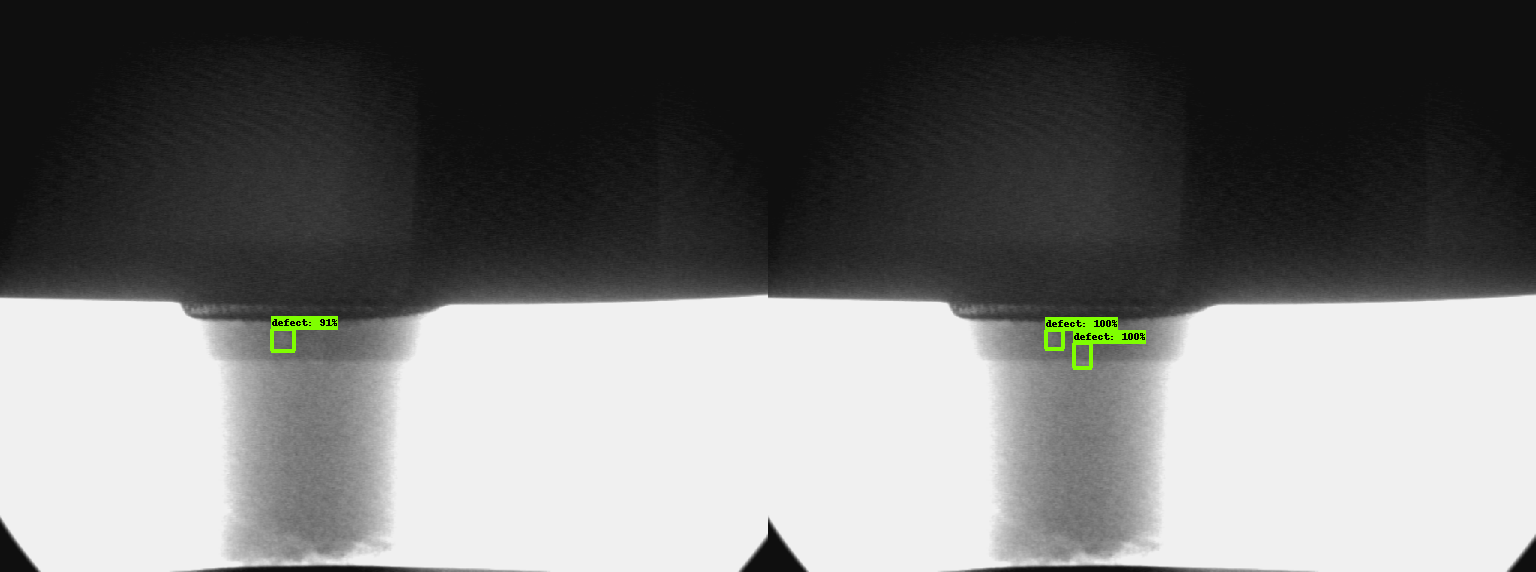}\\
		\includegraphics[scale=0.10]{imageData13} & \includegraphics[scale=0.10]{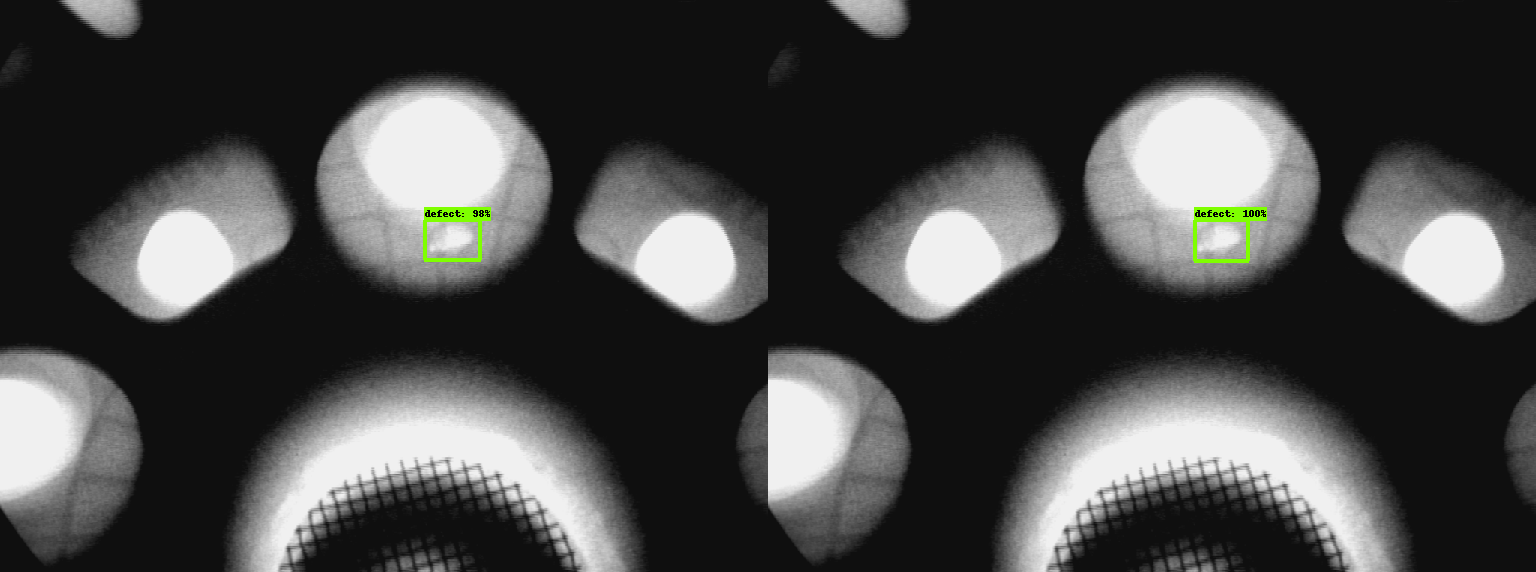}\\
		\includegraphics[scale=0.10]{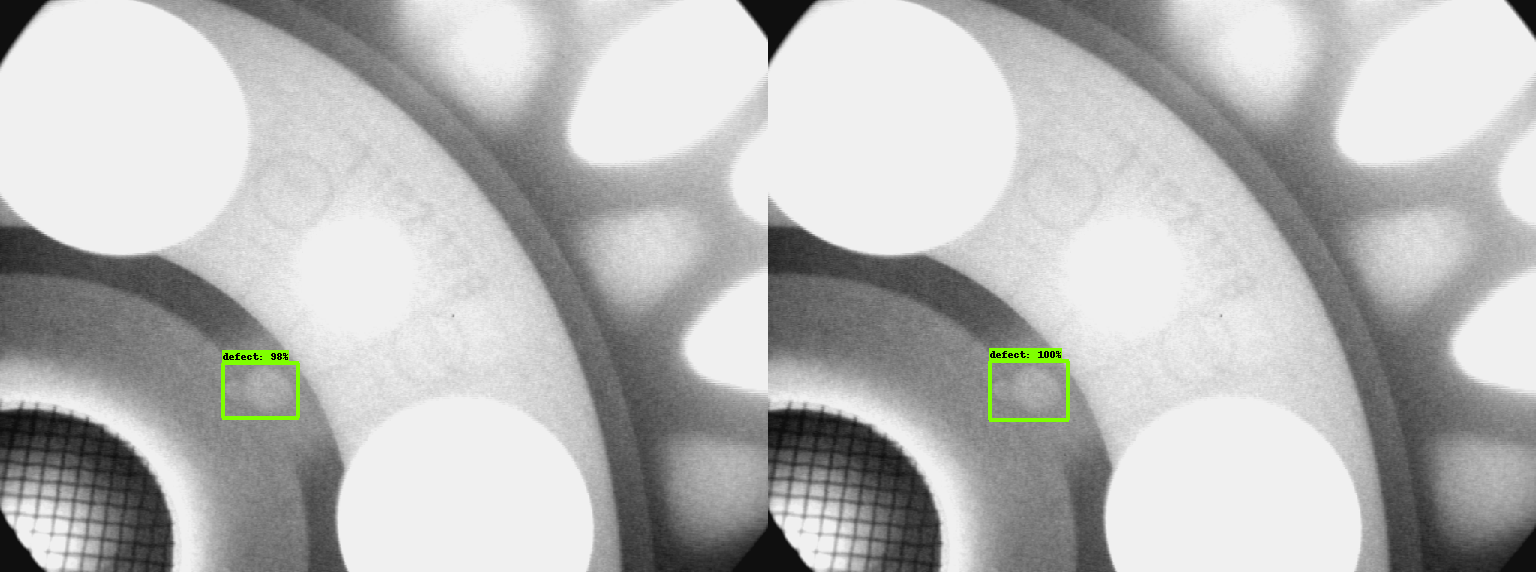} & \includegraphics[scale=0.10]{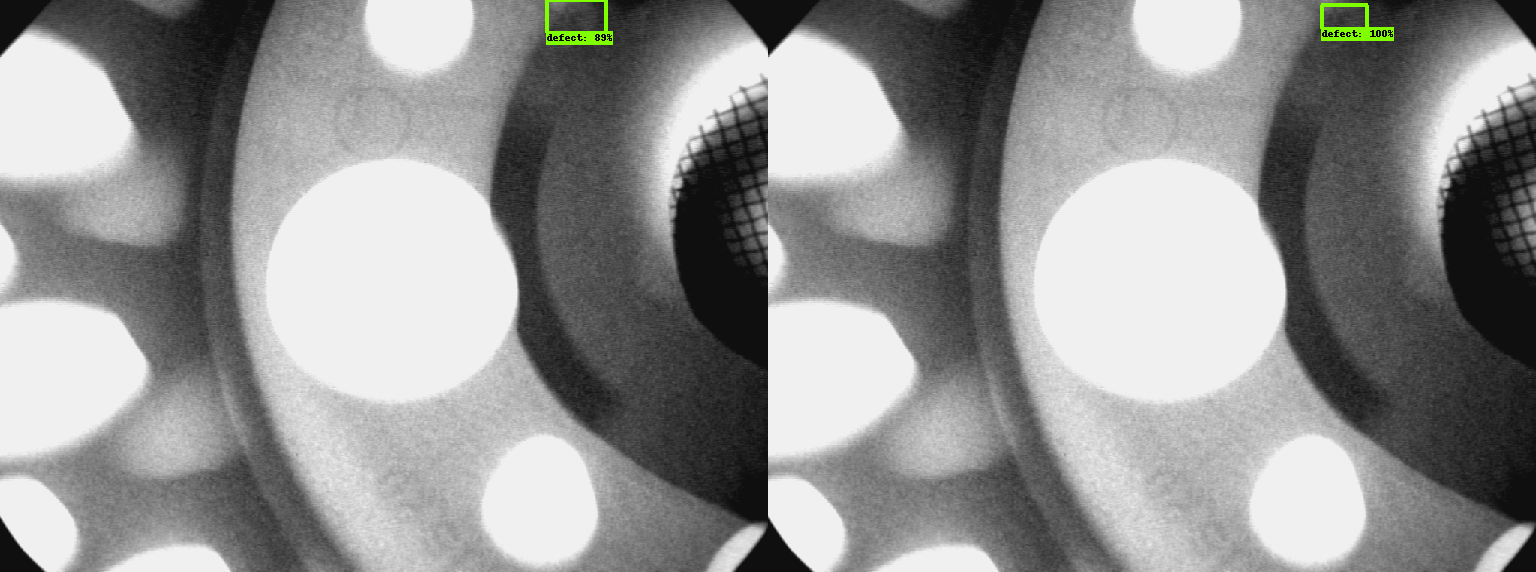}\\
		\includegraphics[scale=0.10]{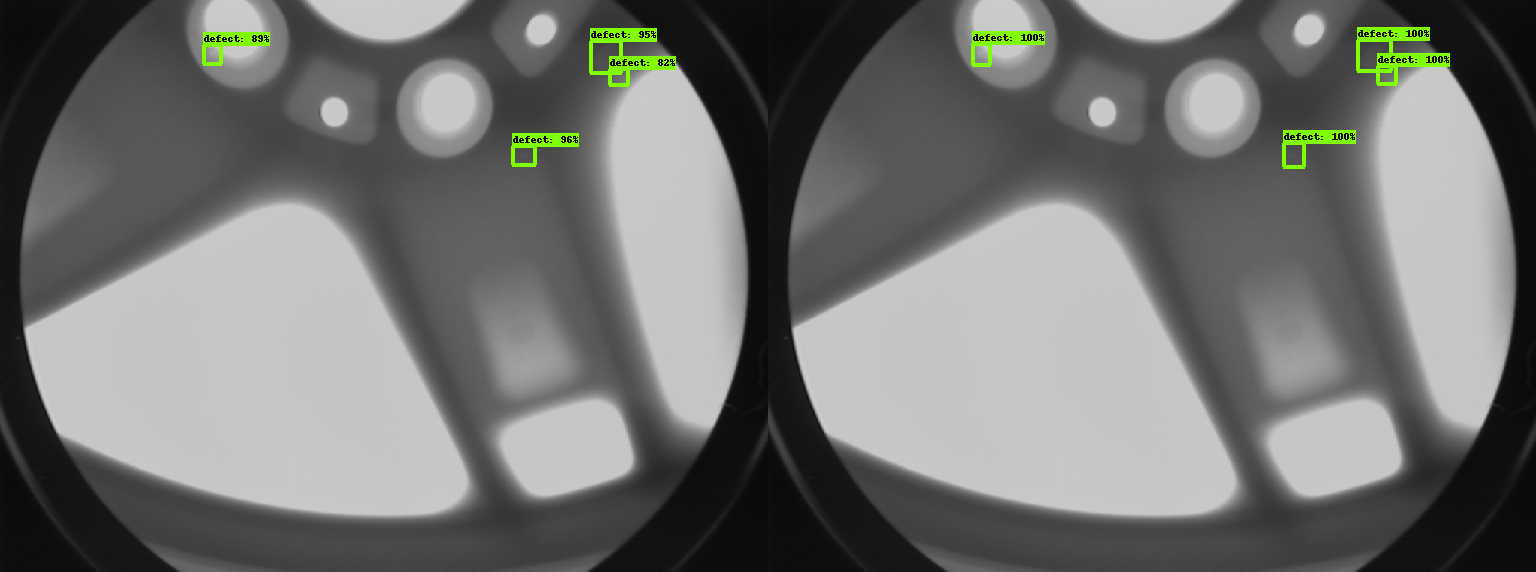} & \includegraphics[scale=0.10]{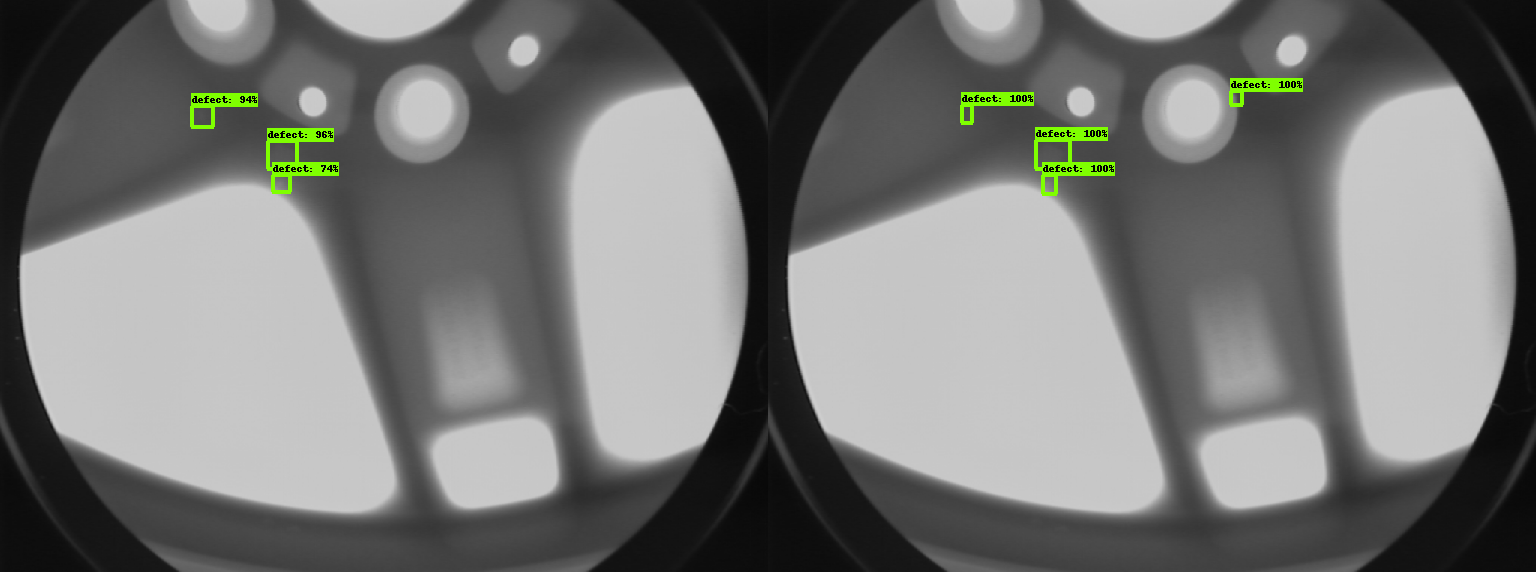}\\
	\end{tabular}
	\caption{Samples of achieved accuracy (II): Model results (left) vs Human inspector (right)}
	\label{fig:inference2}
\end{figure}

On the other hand, increasing batch size usually leads to an increase on model efficiency, and experiment 42 shows that 0.942 mAP@IoU=50\% can be achieved by the RetinaNet model. Table \ref{tab:finalModel} shows all relevant model metrics associated to object detection capability, whilst Figures \ref{fig:inference1} and \ref{fig:inference2}  shows the inference results against GDXray images. On each column, an image is shown with two images: left hand side shows RetinaNet model results and right hand side image shows human inspector Ground Truth defects. In general, a very good correlation can be observed, although in same inference images some defects are missing or False Positive (FP) defects are detected. 

Final model inference time was evaluated on some production X-ray castings images in DICOM format and was found to be less than 400 ms per image, so it can be installed on production facilities with no impact on delivery time.

\begin{table}
	\centering{}%
	\caption{Best model metrics}
	\label{tab:finalModel}
	\begin{tabular}{ll}
		\hline\noalign{\smallskip}
		Parameter & Value\\
		\noalign{\smallskip}\hline\noalign{\smallskip}
		Average Precision  (AP) @[ IoU=0.50:0.95 | area=   all | maxDets=100 ] & 0.580\\ 
		Average Precision  (AP) @[ IoU=0.50      | area=   all | maxDets=100 ] & \textbf{0.942}\\ 
		Average Precision  (AP) @[ IoU=0.75      | area=   all | maxDets=100 ] & 0.612\\ 
		Average Precision  (AP) @[ IoU=0.50:0.95 | area= small | maxDets=100 ] & 0.593\\ 
		Average Precision  (AP) @[ IoU=0.50:0.95 | area=medium | maxDets=100 ] & 0.529\\ 
		Average Precision  (AP) @[ IoU=0.50:0.95 | area= large | maxDets=100 ] & 0.581\\ 
		Average Recall     (AR) @[ IoU=0.50:0.95 | area=   all | maxDets=  1 ] & 0.195\\ 
		Average Recall     (AR) @[ IoU=0.50:0.95 | area=   all | maxDets= 10 ] & 0.396\\ 
		Average Recall     (AR) @[ IoU=0.50:0.95 | area=   all | maxDets=100 ] & 0.649\\ 
		Average Recall     (AR) @[ IoU=0.50:0.95 | area= small | maxDets=100 ] & 0.656\\ 
		Average Recall     (AR) @[ IoU=0.50:0.95 | area=medium | maxDets=100 ] & 0.610\\ 
		Average Recall     (AR) @[ IoU=0.50:0.95 | area= large | maxDets=100 ] & 0.644\\
		\noalign{\smallskip}\hline
	\end{tabular}
\end{table}

\section{Conclusions}
This paper presents an Automatic Defect Recognition system for aluminium casting defects that is around human detection performance whilst exceeds current state of the art object detection models for GDXray dataset. Our model accuracy is 0.942 mAP@IoU=50\%.
The analysis also showed the existence of an optimum number of defects where acceptable accuracy is achieved. Attempts to improve further the model will result on significant effort in industrial organisations as requires a significant increase of production inspector labelled data to obtain a residual improvement on accuracy. It was also highlighted the important of understanding the influence of hyper-parameter optimization, in special, the use of image upscaling to improve detection of the smallest defects and the batch size to achieve a more stable solution.
In spite of the good results, we think this model will not substitute human inspectors but will be a useful production tool to reduce inspection time and improve accuracy, detectability and reliability by removing the subjective interpretation of defects.


%
\section*{Conflict of interest}
The authors declare that they have no conflict of interest.

\newpage
\bibliographystyle{spmpsci}      
\bibliography{Aluminium_Castings_ADR}   

%
%

\end{document}